\documentclass[conference,compsoc]{IEEEtran}
\usepackage{subfigure}
\usepackage{amsmath}
\usepackage{fancyhdr}
\usepackage{multirow}
\usepackage{color}

\definecolor{bluegray}{rgb}{0.4, 0.6, 0.8}
\definecolor{orange1}{rgb}{1.0, 0.49, 0.0}
\definecolor{yellow1}{rgb}{0.99, 0.93, 0.0}
\definecolor{orange2}{rgb}{1.0, 0.75, 0.0}
\definecolor{green2}{rgb}{0.5, 1.0, 0.0}
\definecolor{green1}{rgb}{0.0, 0.5, 0.0}
\definecolor{AEC-green}{rgb}{0.47, 0.87, 0.47}
\definecolor{LA-blue}{rgb}{0.54, 0.81, 0.94}
\definecolor{darkgray}{rgb}{0.66, 0.66, 0.66}
\definecolor{blue1}{rgb}{0.19, 0.55, 0.91}
\definecolor{yellow2}{rgb}{1.0, 0.88, 0.21}
\definecolor{blush}{rgb}{0.87, 0.36, 0.51}
\definecolor{blue2}{rgb}{0.0, 0.75, 1.0}
\definecolor{green2}{rgb}{0.01, 0.75, 0.24}
\definecolor{red2}{rgb}{0.75, 0.0, 0.2}
\definecolor{byzantium}{rgb}{0.44, 0.16, 0.39} 
 	\definecolor{pansypurple}{rgb}{0.47, 0.09, 0.29}
	 	\definecolor{tyrianpurple}{rgb}{0.4, 0.01, 0.24}
 	\definecolor{aureolin}{rgb}{0.99, 0.93, 0.0}
	
\usepackage{graphicx}
\graphicspath{{Fig/}{jpeg/}}
\DeclareGraphicsExtensions{.pdf,.jpeg,.png,.PNG}

\renewcommand{\headrulewidth}{0pt}
\usepackage{lipsum}

\begin{document}

\title{Discovering Interesting Plots in Production Yield Data Analytics}

\author{\IEEEauthorblockN{Matthew Nero, Chuanhe (Jay) Shan, Li-C. Wang}
\IEEEauthorblockA{
University of California, Santa Barbara\\
Santa Barbara, California 93106
}
\and
\IEEEauthorblockN{Nik Sumikawa}
\IEEEauthorblockA{NXP Semiconductor\\
Chandler, AZ 85224
}
}

\maketitle

\begin{abstract}
An analytic process is iterative between two agents, an analyst and an analytic
toolbox. Each iteration comprises three main steps: preparing a dataset, running
an analytic tool, and evaluating the result, where dataset preparation and
result evaluation, conducted by the analyst, are largely domain-knowledge
driven. In this work, the focus is on automating the result evaluation step. 
The underlying problem is to identify plots that are deemed
interesting by an analyst. We propose a methodology to learn such analyst's 
intent based on Generative Adversarial Networks (GANs) and demonstrate its
applications in the context of production yield optimization using
data collected from several product lines. 
\end{abstract}


\thispagestyle{fancy}
\fancyhead{}
\renewcommand{\headrulewidth}{0pt}
\fancyhf{}
\fancyfoot[L]{\large Submission \\ COPYRIGHT LINE}
\fancyfoot[C]{\large INTERNATIONAL TEST CONFERENCE}
\fancyfoot[R]{\large \thepage}


%
\IEEEpeerreviewmaketitle


\section{Introduction}
\vspace{-0.3cm}
\label{sec01}
\label{body:introduction}

\begin{figure}[htb]
\centering
\vspace{-0.2cm}
\includegraphics[width=2.3in]{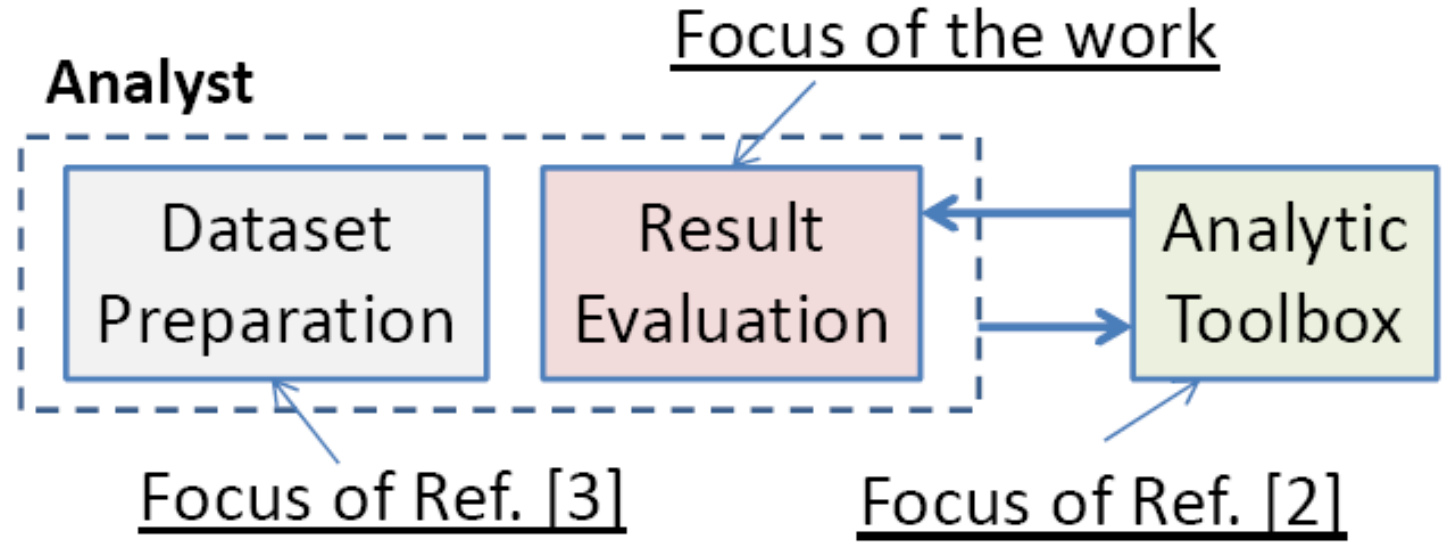}
\vspace{-0.2cm}
\caption{An iterative analytic process comprises three major steps}
\label{fig01}
\end{figure}

Data analytics have been widely applied in design automation and test in recent years.
In certain applications, analytics can be viewed as an iterative search process \cite{wang2016}. 
Such a process comprises three major steps as illustrated in Figure~\ref{fig01}: 
dataset preparation, running an analytic tool, and result evaluation. 
The dataset preparation and result evaluation are largely domain-knowledge driven.
The analytic toolbox comprises statistical or machine learning tools. 

One application that can be seen in terms of this view is production yield optimization. 
Typically, the data to be analyzed are production test data together with  
e-test data characterizing the effects of process on each wafer. The analytics have
two goals: (1) identifying a failure case that impacts the yield, (2) searching to establish
a "high" correlation between one or more e-test parameters and the failing case.
The analytics can be extended to manufacturing data 
if data on manufacturing tools are also available. 

In the context of production yield optimization, for example, the work in \cite{yield2014}
focuses on the analytic toolbox. The study concerns more on what tools are useful or required 
in the application context for resolving a yield issue. Then, the work in \cite{vts2017}
focuses on the dataset preparation. Due to the iterative nature of an analytic process,
dataset preparation can be seen as following a particular flowchart for the search. 
The work \cite{vts2017} concerns how to construct such a flowchart automatically by
learning from an analyst's past experience, i.e. usage log of the analytic toolbox. 

This work focuses on the result evaluation and was motivated by the desire to
automate an entire analytic process as much as possible.
Automation of the toolbox is relatively easy because the input and output
of an analytic tool is usually well defined. Hence, developing an analytic tool is more 
of an algorithmic question. 
Automation of the dataset preparation and result evaluation can be challenging because
these two components are largely domain-knowledge driven. As a result, what it takes
to automate them can be fuzzy or even become an open-ended question.

Automation of the result evaluation component can be considered as a standalone problem and 
essentially means to capture an analyst's intent regarding the interest of a result. 
For example, to search for a high correlation, an analyst constructs a set of datasets 
$D_1, \ldots, D_{k\times n}$ and runs each with a Pearson correlation tool. 
Each dataset comprises $m$ samples. A sample can be defined as 
either a wafer or a lot. Suppose a sample is a lot. There are also choices
to define what each dataset represents. For example, each dataset represents the number
of failing dies from a failing bin. Suppose there are $k$ bins. Further suppose there are 
$n$ e-tests. Then, after pairing each bin with each e-test, there are
$k\times n$ datasets. In each dataset two values are calculated for each sample:
the number of failing dies from the lot and the e-test value. The e-test value, for example,
can be the average of measured values across all wafers in a given lot. 

\begin{figure}[htb]
\centering
\vspace{-0.2cm}
\includegraphics[width=2.8in]{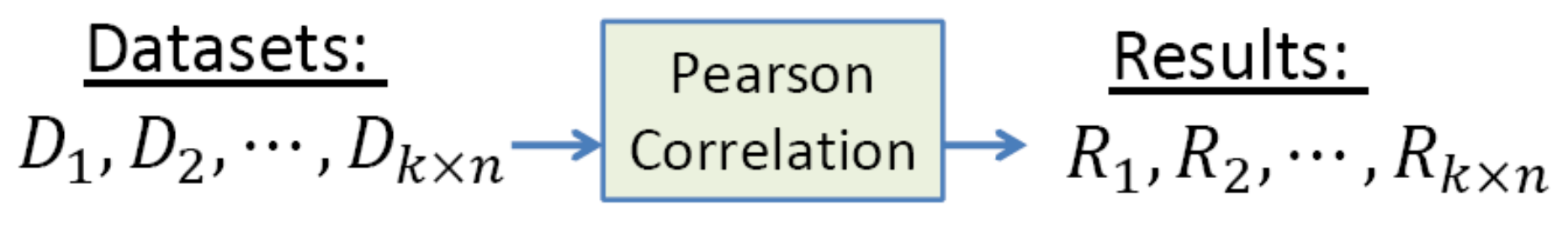}
\vspace{-0.2cm}
\caption{Running a correlation tool over a set of datasets}
\label{fig02}
\end{figure}

Running the tool leads to results $R_1, \ldots, R_{k\times n}$ as illustrated
in Figure~\ref{fig02}. Each $R_i$ can contain two pieces of result, a correlation
number and a correlation plot. In this search, the analyst's intent is to find
a "high correlation." Usually, a script can be written to check the results to
see if any correlation number is greater than, say 0.8. This approach essentially
is using a {\it fixed rule} to capture the intent. 

If the analyst's intent can be captured completely with a fixed rule, then 
there is no problem to automate it. The issue is that in many situations,
completely capturing an intent is not so easy. For example, when a rule is fixed
to search for a correlation number greater than or equal to 0.8, it would overlook 
any result with the number less than 0.8. More importantly, a low correlation
number does not imply the corresponding plot is not interesting to the analyst. 

\begin{figure}[htb]
\centering
\vspace{-0.2cm}
\includegraphics[width=3in]{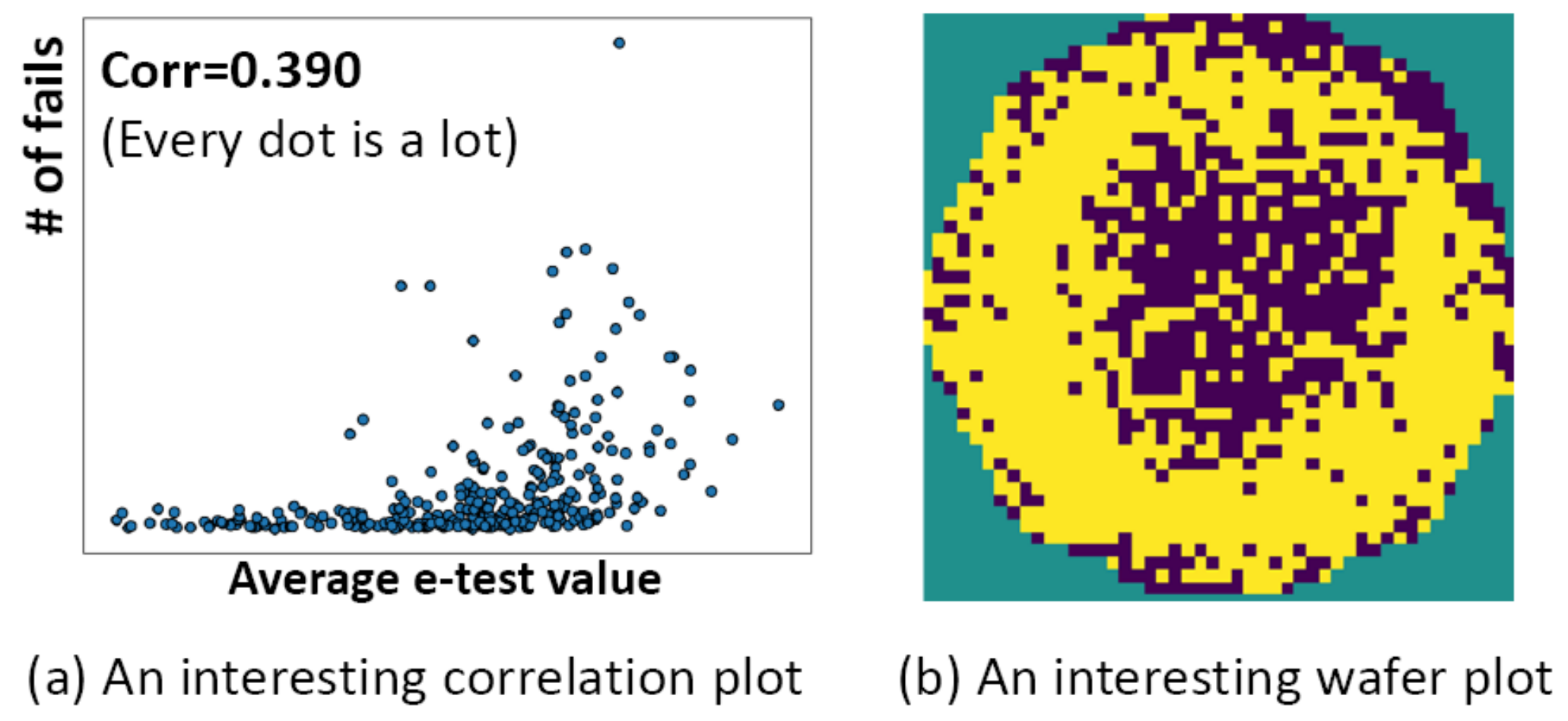}
\vspace{-0.2cm}
\caption{Interesting plots not easily described with a fixed rule}
\label{fig03}
\vspace{-0.1cm}
\end{figure}

For example, Figure~\ref{fig03}-(a) shows a correlation plot that is interesting
(the interest will be explained in detail later). While the correlation number
is low, the plot shows that increasing e-test value tends to have more fails. 
Consequently, adjusting the process to reduce the particular parameter
value can be a choice for improving the yield. 

Figure~\ref{fig03}-(b) shows another example of interesting plot. The plot
shows a wafer and the location of its {\it passing} dies marked as 
{\color{tyrianpurple} dark-purple dots}. 
This plot is interesting not only because it shows a cluster of passing dies but
also there seems to be three concentrating sub-clusters on the center
(the reason will be explained later). 

If an analyst knew in advance to specifically look for those particular 
plots as shown in Figure~\ref{fig03}, then presumably the analyst could write a rule to
describe what to look for. However, seeing examples of 
interesting plots does not always imply it is easy for the analyst to write a rule
to capture the intent. For example, the work in \cite{nik2017} develops
a non-trivial rule in order to automatically recognize wafers with certain 
class of clustered fails. 

Depending on the experience of an analyst, the analyst might not know
what interesting plots to look for in advance or not know how to write a rule
to capture a class of interesting plots. In practice, it can be
difficult to manually enumerate all classes of interesting plots
in advance and describe each with a fixed rule. If this difficulty cannot
be overcome in practice for an analytic process, then using fixed rules
to implement the result evaluation component can become
an ineffective approach. 

Motivated by the need to overcome the difficulty, this work pursues an
alternative approach than using a fixed-rule. The assumption is that
it is easy for a person to judge if a plot, when presented to the
person, is interesting or not. Based on this assumption, a machine
learning model can be trained to recognize a particular class of plots
(interesting or non-interesting). 
Such a model can be used to capture an analyst's intent based on 
learning from the example plots. 

\fancyhf{}
\fancyfoot[L]{\large Submission}
\fancyfoot[C]{\large INTERNATIONAL TEST CONFERENCE}
\fancyfoot[R]{\large \thepage}

Developing a plot recognizer can be seen as an unsupervised learning
problem. The training data can comprise only one particular class of
plots. Recent advances in unsupervised
learning based on Generative Adversarial Networks (GANs) \cite{GAN2014} 
provide a good underlying technology to implement such a recognizer. 
In this work, we therefore investigate how to build a plot recognizer
using GANs. With such recognizers, we develop a methodology to apply
them in practice. In particular, the methodology is applied to analyze
production data from several product lines. Its usefulness
will be explained through several findings that led to an 
improvement on the yield. 

Note that the recognizer-based methodology is not proposed as a replacement
for the fixed-rule approach. It is an alternative that can be applied when
developing a fixed-rule is practically difficult. The degree of this
difficulty, however, can vary from analyst to analyst. Hence, one 
cannot say that it is impossible for any fixed-rule approach
to capture what being captured by a recognizer. Can a well-trained very
intelligent analyst to develop a sophisticated rule to also capture
an intent captured by a recognizer? Possibly. But this is not the question studied
in this work. 

It is also interesting to note that very often the end result of 
an analytic process shown in Figure~\ref{fig01} is a PowerPoint presentation 
containing slides of interesting plots. 
The proposed methodology is a way to improve the efficiency for discovering 
those interesting plots. Applying the
methodology results in three sets of plots: non-interesting plots,
known interesting plots, and unrecognized plots. The idea is that the number
of known interesting plots and unrecognized plots are small enough for the analyst
to inspect carefully, and select some to be included in the presentation. 

The rest of the paper is organized as the following.
Section~\ref{sec02} introduces the GANs approach for learning a plot 
recognizer.
Section~\ref{sec03} describes two product lines and their 
data used in most of the study. 
Section~\ref{sec04} discusses the three types of analytics considered
and a methodology to develop the plot recognizers.  
Section~\ref{sec05} focuses on discovering
interesting wafer plots and discusses an application scenario. 
Section~\ref{sec06} focuses on
correlation plots and discusses another application scenario. 
Section~\ref{sec07} focuses on box plots and 
discusses two other application scenarios.  
Section~\ref{sec08} concludes. 

\vspace{-0.1cm}
\section{Developing a GANs-based Recognizer
\protect\footnote{Please note that this part is not intended to reiterate the materials in 
\cite{GAN2014}\cite{CNNGAN2016}\cite{GAN2016}. Instead, the focus is
more on the key aspects to which attention should be given for 
implementing a GANs-based plot recognizer in practice. 
We apologize for not including all the detail regarding GANs.
}
}
\label{sec02}
\vspace{-0.2cm}

\begin{figure}[htb]
\centering
\vspace{-0.2cm}
\includegraphics[width=2.6in]{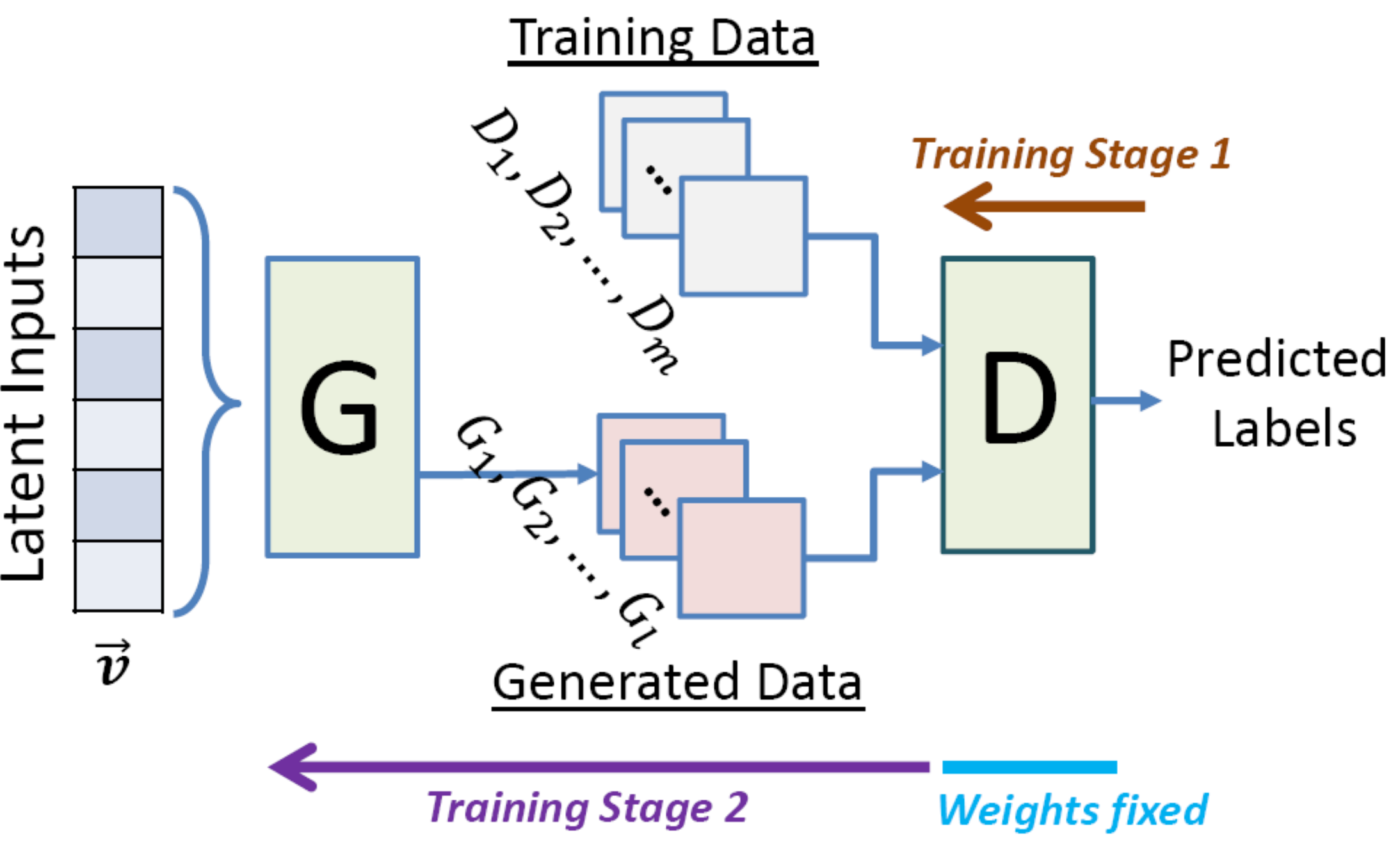}
\vspace{-0.2cm}
\caption{Illustration of GANs and their training}
\label{fig04}
\vspace{-0.2cm}
\end{figure}

Generative Adversarial Networks (GANs) \cite{GAN2014} are methods to learn a generative model. 
Given a dataset, a generative model is a model that can synthesize new samples
similar to the training samples. A GANs architecture consists of two neural networks.
The generator network {\bf G} is trained to produce the samples. The discriminator
network {\bf D} is trained to differentiate the training samples from the samples
produced by the generator. Figure~\ref{fig04} illustrates the design of GANs. While the main 
goal of GANs is to learn the generator, after the training, the discriminator can be
used as a recognizer for future samples similar to the training samples. 
Hence, in this work, our interest is in training a discriminator to be a
recognizer for a class of plots. 

To train a recognizer, a class of plots are used. Suppose there are $m$ plots
and denoted as $D_1, \ldots, D_m$. These are our training data. 
Without loss of generality, assume each plot is a square image. 
For training, the generator produces some $l$ images, denoted as
$G_1, \ldots, G_l$. Each generated image is produced according to a
random vector $\vec{v}$. Each variable of $\vec{v}$ can be thought of
as a {\it latent input}. These variables define a {\it latent space}
where each vector in this space represents an image produced by
the generator. 

The training process is iterative. Each iteration has two stages and
each stage of training can use the common stochastic gradient descent (SGD) approach.
In each iteration, two classes of samples $D_1, \ldots, D_m$ and
$G_1, \ldots, G_l$ are used. From iteration to iteration, the 
samples $D_1, \ldots, D_m$ remain the same, but $G_1, \ldots, G_l$
are re-produced by the generator for each iteration based on the weights 
learned in the previous iteration. 

In the first stage of training, the goal is to learn the weights in the
{\bf D} network in order to separate $D_1, \ldots, D_m$ from 
$G_1, \ldots, G_l$ as much as possible. During back propagation,
the gradients are computed backward from the output of {\bf D}
to its inputs. In the second stage, weights in {\bf D} are fixed. 
SGD is applied to learning the weights in {\bf G}. The gradients 
calculated on inputs of {\bf D} are further back propagated to the
inputs of {\bf G}. The optimization objective is to have {\bf G} 
adjust the generated samples such that their output labels by
{\bf D} are as close as possible to the output labels of 
the training samples $D_1, \ldots, D_m$. 

The idea of training {\bf D} and {\bf G} can be thought of as
playing a game \cite{GAN2014} where the {\bf D} network learns to beat the
{\bf G} network by discriminating the samples generated by {\bf G}
from the training samples $D_1, \ldots, D_m$. On the other hand, the {\bf G}
network learns to generate samples to fool the discriminator {\bf D}
as much as possible. Over iterations, the generated samples become more
like the training samples and it becomes harder for {\bf D} to separate them.

\vspace{-0.1cm}
\subsection{The CNN architectures}
\label{sec02.1}
\vspace{-0.3cm}

For computer vision applications, convolutional neural networks (CNNs) have 
shown remarkable performance in the context of supervised learning in recent 
years. Using CNNs for unsupervised learning had received less attention until
the GANs approach was proposed. In this work, our implementation of GANs
is based on two deep CNNs, following the ideas proposed in \cite{CNNGAN2016}
which suggests a set of constraints on the architectural topology of 
Convolutional GANs to make them stable to train. 

\begin{figure}[htb]
\centering
\vspace{-0.2cm}
\includegraphics[width=3.3in]{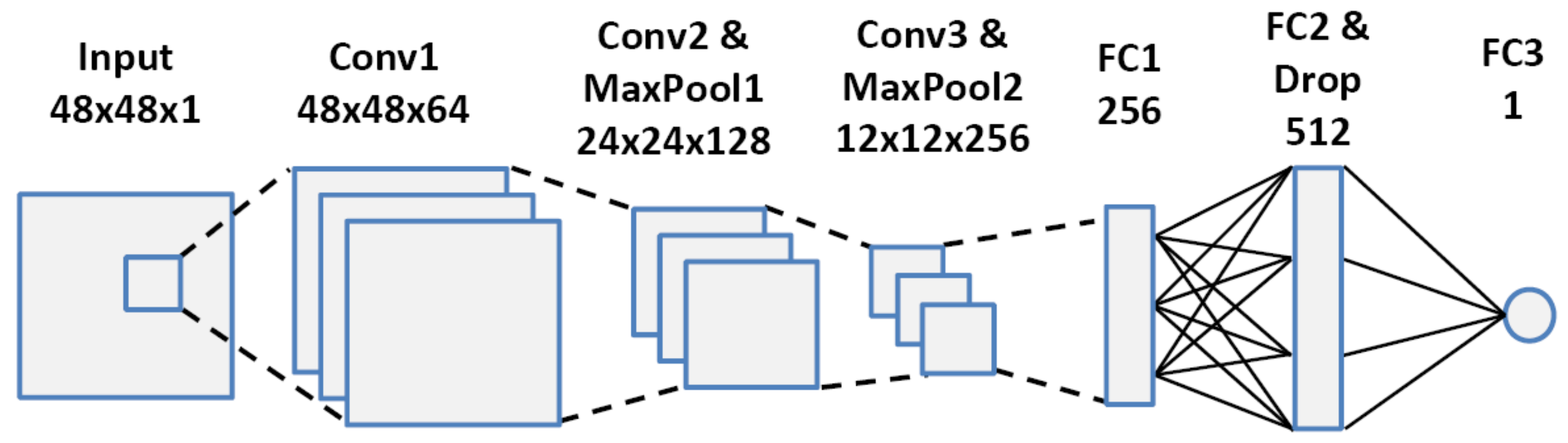}
\vspace{-0.2cm}
\caption{CNN architecture for the discriminator}
\label{fig05}
\vspace{-0.2cm}
\end{figure}

Figure~\ref{fig05} shows our CNN architecture for the discriminator. This
architecture is used for training all types of plot recognizers studied 
in this work. The leftmost block shows our input assumption. Each input
is an image with 48-by-48 pixels. Each pixel has three values: -1, 0, and +1. 
These three values indicate negative color, no color, and positive color,
respectively. Before a plot can be used as an input sample to this CNN,
preprocessing is required to convert the plot into this representation.  

The CNN has three convolutional layers (Conv1 to Conv3) where after
Conv2 and Conv3, there is a Max pooling layer denoted as
MaxPool1 and MaxPool2, respectively. After the MaxPool2 layer, there
are three fully-connected layers (FC1 to FC3). 
The size and number of channels after each layer are denoted in
the figure. For example, after Conv1 the image is transformed from 1 channel 
of 48$\times$48 to 64 channels of 48$\times$48, using
64 2$\times$2 filters (In our CNNs, the filter size is always 2$\times$2). 
After Conv2/MaxPool1, the image size is reduced to 24$\times$24
with 128 channels. 

The fully-connected layer FC1 has 256 perceptrons (artificial neurons)
each receiving inputs from all the 12$\times$12$\times$256 perceptrons
in the previous layer. The FC2 has 512 perceptrons. The last layer
FC3 has one perceptron which outputs a classification probability. 
As suggested in \cite{CNNGAN2016}, Leaky ReLU is used as the
activation function for all perceptons in the CNN. Each perceptron
also includes a bias parameter. The total number of parameters
(weights) in the CNN is 9,734,592.

\begin{figure}[htb]
\centering
\vspace{-0.2cm}
\includegraphics[width=3.3in]{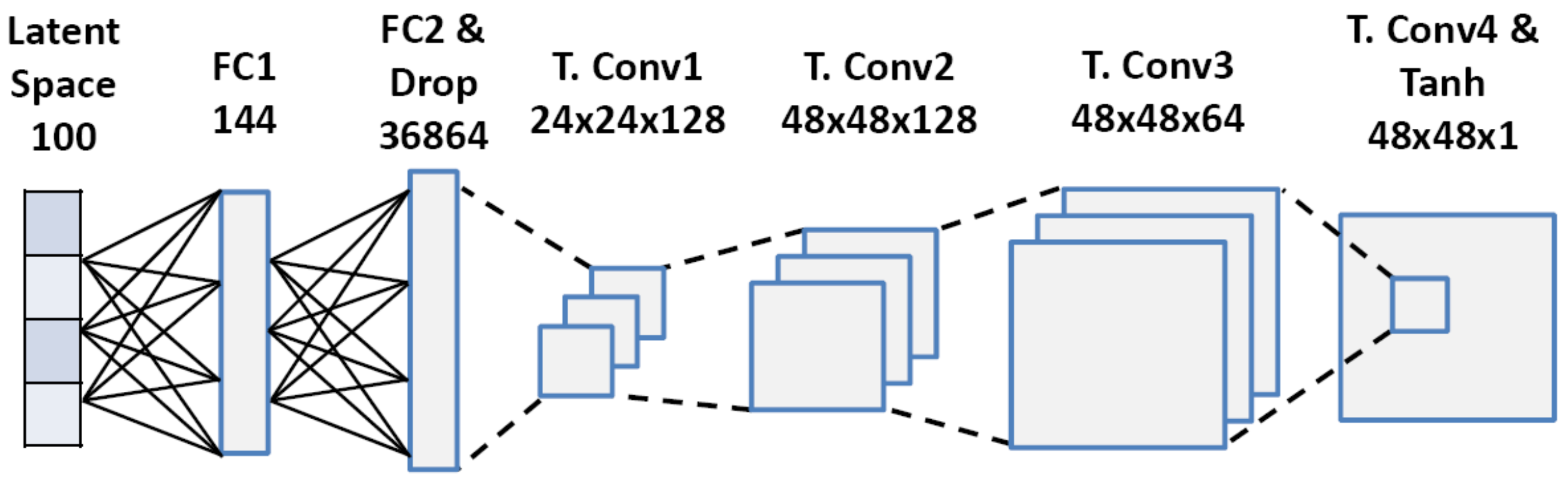}
\vspace{-0.2cm}
\caption{CNN architecture for the generator}
\label{fig06}
\vspace{-0.2cm}
\end{figure}

Figure~\ref{fig06} shows the CNN architecture for the generator. 
There are two fully-connected layers, FC1 and FC2, and
four transposed convolutional layers, T.Conv1 to T.Conv4. 
Like the discriminator CNN, Leaky RuLU and bias parameter are used for 
all perceptrons. The generator CNN can be thought as the reverse
of the discriminator CNN. For the generator CNN, the number of parameters
(weights) is 24,333,009. Together, the total number of parameters
to be trained in the GANs is 34,067,601.

\vspace{-0.1cm}
\subsection{Implementation detail}
\label{sec02.2}
\vspace{-0.3cm}

For training GANs, attention is required to ensure two aspects:
the convergence of the training iteration and the output quality of 
both CNNs. The work in \cite{CNNGAN2016} suggests several architectural
guidelines to improve quality. Among them, we found that the performance
of the CNNs are sensitive to whether or not we chose to use (1)
the Batchnorm in both generator and discriminator CNNs, and (2) the
Leaky ReLU activation function for all perceptrons. 
For convergence, we found that the feature matching technique proposed
in \cite{GAN2016} is crucial. Otherwise, it is difficult for the
training to converge. 

Although the Leaky ReLU is used for all perceptrons, in the discriminator
CNN, the Sigmoid function is used to convert the output of the last perceptron 
into a value between 0 and 1. Similarly, a Hyperbolic Tangent function is used in
the generator CNN for adjusting the output value. 

The CNNs are implemented with Google TensorFlow \cite{tensorflow2015-whitepaper}
and run with the nVidia GTX 980Ti GPU. The optimizer used for the training is
ADAM optimizer \cite{Adam2017}. We had tried others such as regular SGD and AdaDelta but they
did not allow convergence as fast as the ADAM optimizer. 

Two things to note regarding the CNNs in Figure~\ref{fig05} and Figure~\ref{fig06}
are: (1) It is important to include the fully-connected layers for training
a good-quality recognizer (the discriminator). (2) It is important to implement
a Dropout strategy in FC2 \cite{Dropout2014}, the largest layer in each CNN.

\vspace{-0.1cm}
\subsection{An example recognizer for a wafer pattern}
\label{sec02.3}
\vspace{-0.3cm}

To train our GANs,  we need a dataset divided into a training dataset
and a validation dataset. Because it is an unsupervised learning, 
the validation dataset alone cannot fully determine the stopping point.
The validation dataset is used to ensure the discriminator does not
over-fit the samples in the training dataset, by ensuring that all
samples in the validation set are also classified correctly. 
In our experiments, the stopping point in the training is assisted
by inspecting the samples generated by the generator. If these samples
show features similar to the training samples, then we stop. If not,
the training is resumed for more iterations. 

Because our focus is on the discriminator (used as a plot recognizer), 
we concern more about the quality of the discriminator than the quality 
of the generator. If the latter is our concern, we might need to train 
with more iterations
until the generator is capable of producing plots close to the training
samples. This in turn might require additional techniques
in the implementation to ensure convergence. 

\begin{figure}[htb]
\centering
\vspace{-0.2cm}
\includegraphics[width=3.3in]{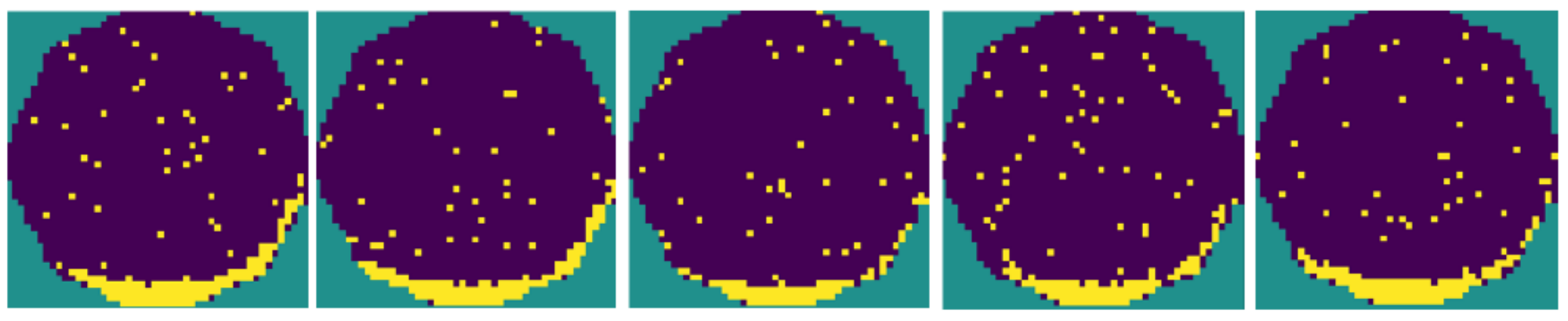}
\vspace{-0.2cm}
\caption{Five training samples}
\vspace{-0.2cm}
\label{fig07}
\end{figure}

What we found is that the GANs usually do not require a large number of
samples to train if those samples share some common features. 
To illustrate this, Figure~\ref{fig07} shows five
training samples used for training a recognizer for this class of wafer
pattern. The {\color{yellow2} yellow dots} represents failing dies. 
To enhance the training dataset, each sample is incrementally rotated 
to produce 12 samples in total. Then, overall we have 60 samples for training.

\begin{figure}[htb]
\centering
\vspace{-0.2cm}
\includegraphics[width=3.3in]{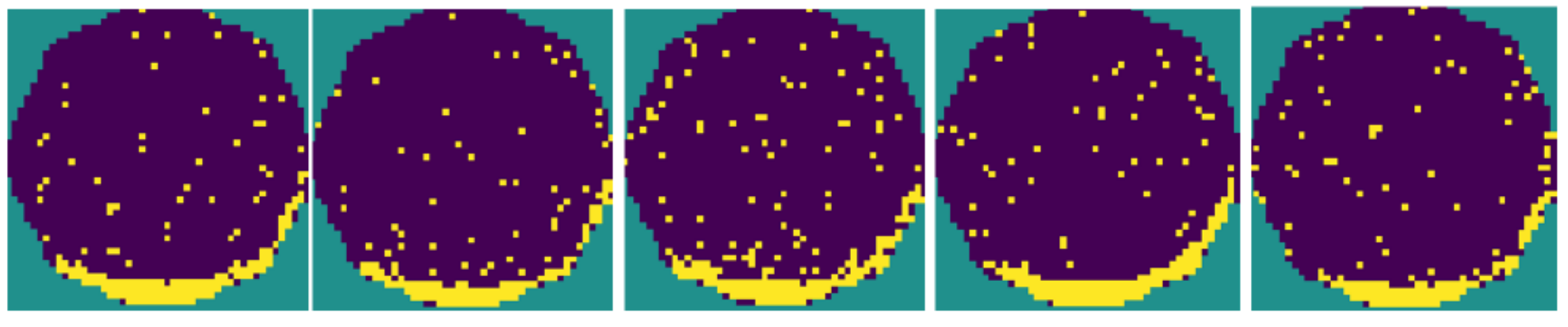}
\vspace{-0.2cm}
\caption{Validation samples}
\label{fig07a}
\vspace{-0.2cm}
\end{figure}

Figure~\ref{fig07a} shows the five samples used for validation.
Similarly, each is rotated to produce 12 samples with a total
of 60 validation samples. Note that these samples look alike because
these are wafers from the same lot. 

\begin{figure}[htb]
\centering
\vspace{-0.2cm}
\includegraphics[width=3.3in]{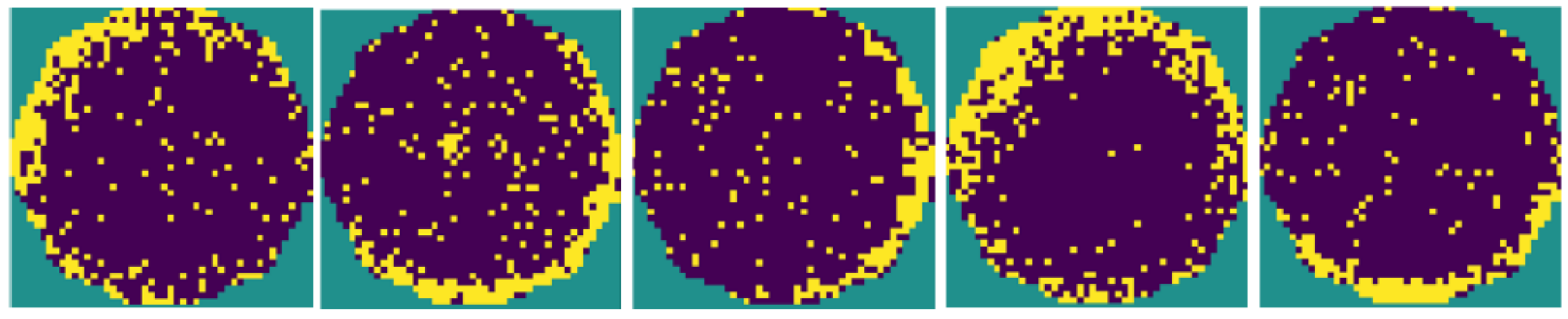}
\vspace{-0.2cm}
\caption{Five wafers among the 25 recognized wafers}
\label{fig09}
\end{figure}

The training took about 2 hours with a total of 3650 iterations. 
After the training, the discriminator (treated as our plot recognizer) 
is used to recognize similar wafer patterns on 8300 other wafers. 
The recognizer recognizes 25 wafers and some are shown in Figure~\ref{fig09}.
On these samples, we see that they all show up with a clear edge failing
pattern. 

\begin{figure}[htb]
\centering
\vspace{-0.2cm}
\includegraphics[width=3.3in]{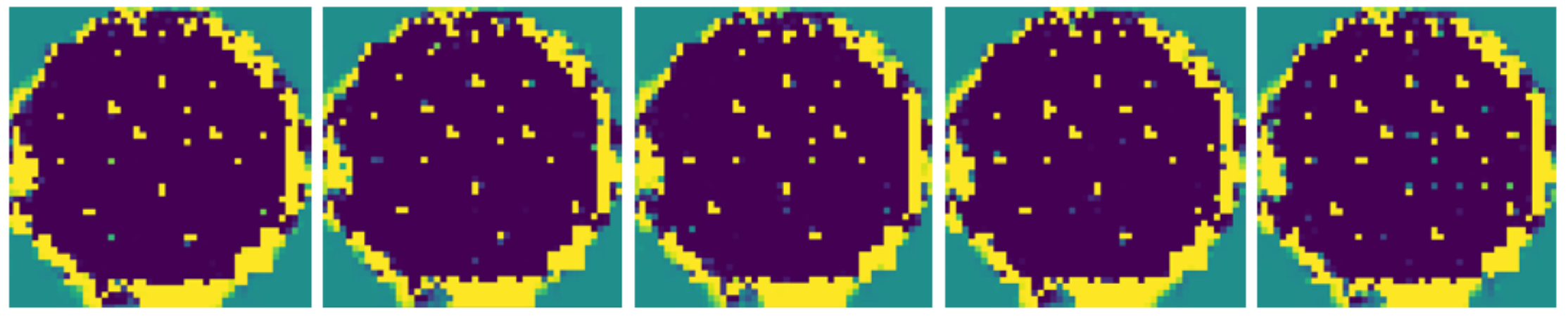}
\vspace{-0.2cm}
\caption{Wafer plots produced by the generator}
\vspace{-0.2cm}
\label{fig10}
\end{figure}

Because the samples generated by the generator are inspected to determine
the stopping point, it would be interesting to show the wafer plots produced
by the generator. Figure~\ref{fig10} shows five such wafer plots (by giving
the generator five random inputs). 
It can be seen that the generated plots do not look the same as the
original plots shown in Figure~\ref{fig07}. However, the feature of
having many fails on the edge are also present in these generated plots.

\vspace{-0.2cm}
\subsection{Generality of the plot recognizer}
\label{sec02.4}
\vspace{-0.3cm}

The wafers used in the above experiment each has about 2100 dies. 
Recall from Figure~\ref{fig05} that input images are in size
of 48$\times$48 pixels. One interesting question to ask would be
how the recognizer performs on those wafer plots from other product lines
where each plot is based on more or has less number of dies. 

\begin{figure}[htb]
\centering
\vspace{-0.2cm}
\includegraphics[width=3in]{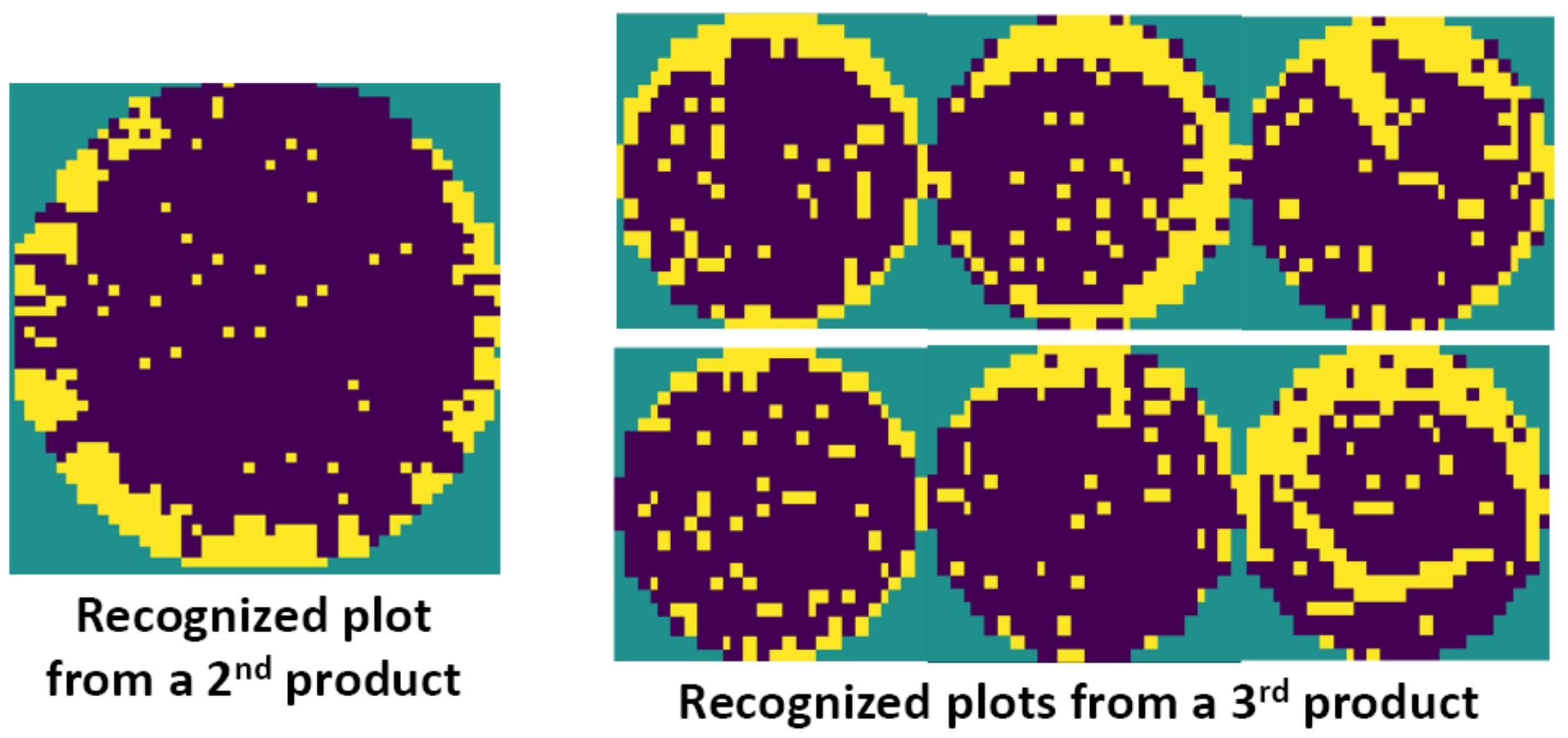}
\vspace{-0.2cm}
\caption{Recognized plots from the 2nd and 3rd product lines}
\vspace{-0.2cm}
\label{fig11}
\end{figure}

To answer this question, Figure~\ref{fig11} shows the result of
applying the edge pattern recognizer on a 2nd product line. The recognizer
was applied to scan 2011 wafer plots and found only 1 recognized plot
as shown in the figure. Each wafer for this product line has about 4500
dies, more than twice as many as that in the first product line used
for the experiment above. It is interesting to see that the recognized
plot also shows a clear edge failing pattern. 

Then, the recognizer was applied to a 3rd product line to scan 7052
wafer plots and found 24 recognized wafer plots. Some examples are
also shown in Figure~\ref{fig11}. For this product line, each wafer has
about 440 dies, much less than that in the first product line. 
However, each recognized wafer plots also show a clear edge failing
pattern. 

From the results shown above across three product lines, it is interesting
to see that the plot recognizer, trained with a small set of rather similar
edge failing patterns as shown in Figure~\ref{fig07}, is able to generalize
the learning and recognize other edge failing patterns even with some
noise in the pattern (Figure~\ref{fig11}). These results show that one might
not need to re-train a recognizer for every product line even though their
numbers of dies per wafer are different. These results also indicate that
one can train a recognizer with some higher-level "intent." For example,
in the above, our intent is to capture an "edge failing pattern."

\vspace{-0.2cm}
\section{Production Data and Analytics}
\label{sec03}
\vspace{-0.2cm}

\begin{figure}[htb]
\centering
\vspace{-0.2cm}
\includegraphics[width=3.3in]{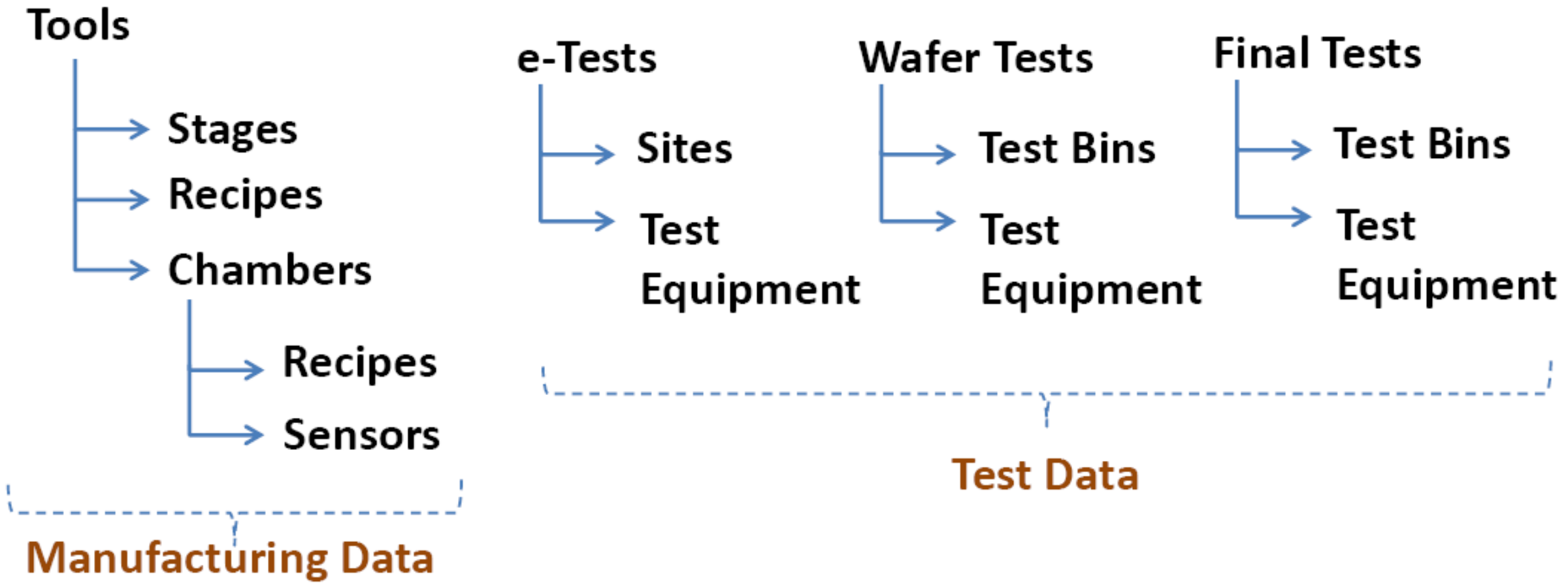}
\vspace{-0.2cm}
\caption{Categories of production data for the analytics}
\label{fig13}
\end{figure}

In the rest of the paper, the results are presented based on data
collected from two product lines, one for the medical market (call it
product M) and the other for the automotive market (call it product A). 
Figure~\ref{fig13} illustrates four categories of production data
used in our study. Production data refers to all the data, which
can further be divided into manufacturing data and test data. 

In production, every lot goes through a sequence of {\it tools} 
(manufacturing equipments). There could be more than hundreds of 
tools involved. Each tool may process one of more {\it stages}. 
A stage has its own recipe name. 
A stage can be carried out by tools arranged in parallel. Then, two
lots may go through two different tools. 
A tool can have multiple {\it chambers}. The chambers can be in
sequence and/or in parallel. Hence, two wafers may go through two
different chambers if they are arranged in parallel. Each chamber
has its own recipe name. Recipe name can change over time. 
Moreover, many sensors are used to measure
properties of a chamber, such as temperature, pressure, etc. Sensor
data are associated with the chambers. The sensor data points are 
incremented over a small time interval.  

The manufacturing data is organized at three levels. The first level
is organized by lots, showing which tools a lot goes through and when.
The second level is organized by wafers, showing which chambers a wafer 
goes through and when. The third level is organized by chambers, showing 
the sensor data in terms of waveform signals over time. 

The test data is organized in three groups:
e-tests, wafer tests, and final tests. 
For e-tests the data contains information regarding the measurement
sites and also the equipment used to measure them. 
For wafer and final tests, the data contains a list of failing bins
including the test a die fails on. The data also contains information
regarding the test equipment. 
The test data are indexed so that one can organize the information
in terms of die, wafer, or lot for an analytic task. It is possible
to arrange the data over a time index so that a particular result can be 
tracked over time. 

\begin{table}[htb]
\vspace{-0.2cm}
\centering
\setlength{\tabcolsep}{4pt}
\caption{Some information about the data collected}
\vspace{-0.3cm}
\begin{tabular}{|c|cc|ccc|} \hline
Product        & wafers & e-tests & probe tests & final tests & total \# of bins \\ \hline
M    & 8300 & 501 & 140  & 400  & 289  \\
A    & 7052 & 596 & 2367 & $>$10k & 132  \\ \hline
\end{tabular}
\label{table1}
\vspace{-0.2cm}
\end{table}

Table~\ref{table1} shows some some numbers regarding the test data. 
Our experiments use mostly the test data
and one experiment uses part of the manufacturing data. 

\vspace{-0.1cm}
\section{Types of Analytics (Plots)}
\label{sec04}
\vspace{-0.2cm}

In production yield analytics, an analyst can apply many different types
of analytics to examine the data described above. In this work, we focus
our discussion on three basic types of analytics generating three types
of plots: wafer plots, correlation plots, and box plots. 
We use these three types of plots to illustrate our recognizer-based
methodology which can be applied to other types of plots as well. 

In section~\ref{sec02}, many examples of wafer plots are already shown. 
A typical wafer plot has two colors, one to show the failing die
locations and the other to show the passing die locations. 
A wafer plot can be based on all the failing dies, or only those
collected in a particular test bin, due to a particular test, or
at a particular test value range. Wafer plots are useful in production
yield optimization because a plot may show an unusual failing pattern.
If this pattern persists over time, it may become a strong indicator for
an issue with a particular process tool, stage, or chamber. 

Correlation plots are often used to assist in correlation analysis. 
For a plot, the analyst decides what variable to use for the x-axis
and what variable to use for the y-axis. For example, x-axis variable
can be an e-test and y-axis variable can be the yield based on a
selected test bin. On a plot, typically the analyst is interested in
discovering a "trend," for example "large y values tend to imply
large x values." 
Correlation plots are often used to relate a failing case to a 
process parameter. 

\begin{figure}[htb]
\centering
\vspace{-0.2cm}
\includegraphics[width=2.6in]{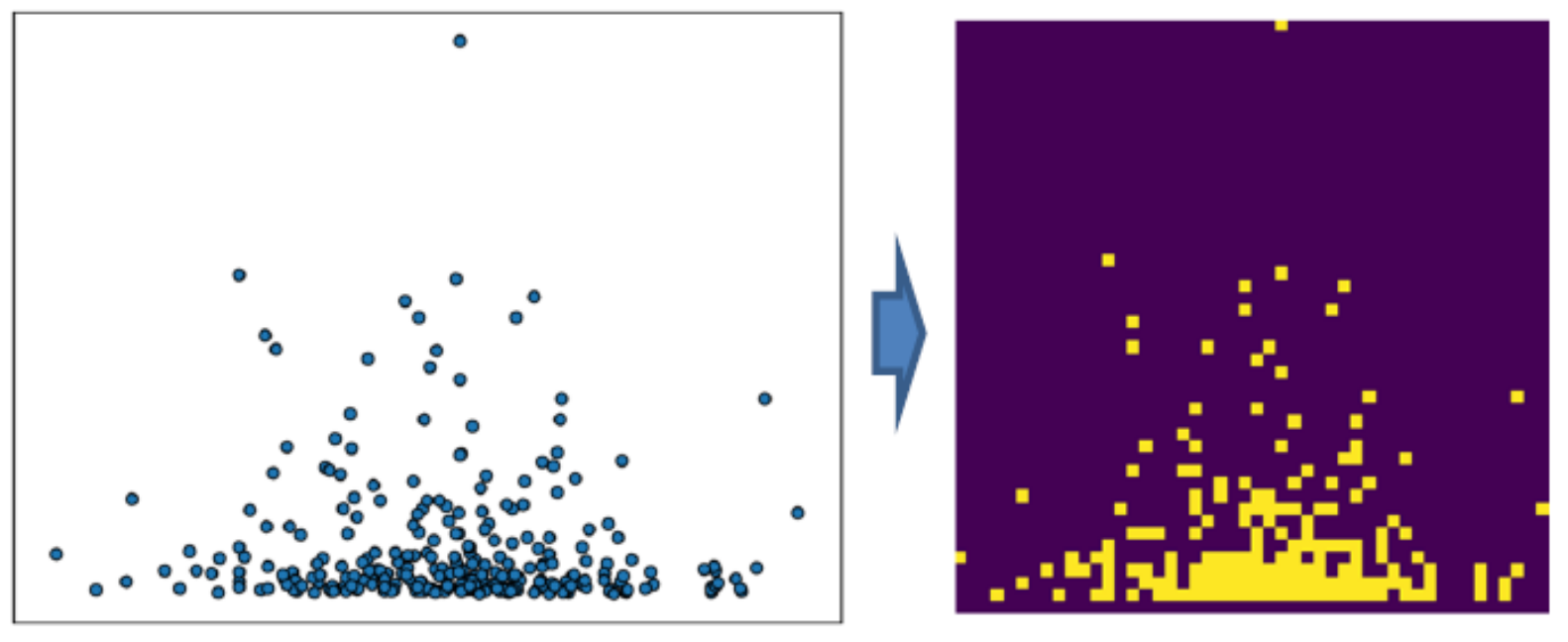}
\vspace{-0.2cm}
\caption{A correlation plot and its transformed image}
\label{fig14}
\vspace{-0.4cm}
\end{figure}

Recall from Figure~\ref{fig05} that our plot recognizer assumes the 
input image size is 48$\times$48. Hence, internally a recognizer
sees the plot differently from that seen by an analyst. 
For example, Figure~\ref{fig14} shows an example where the left correlation
plot is what is being seen by a person and the right 
plot represents what is being seen by the recognizer. 

\begin{figure}[htb]
\centering
\vspace{-0.2cm}
\includegraphics[width=3.3in]{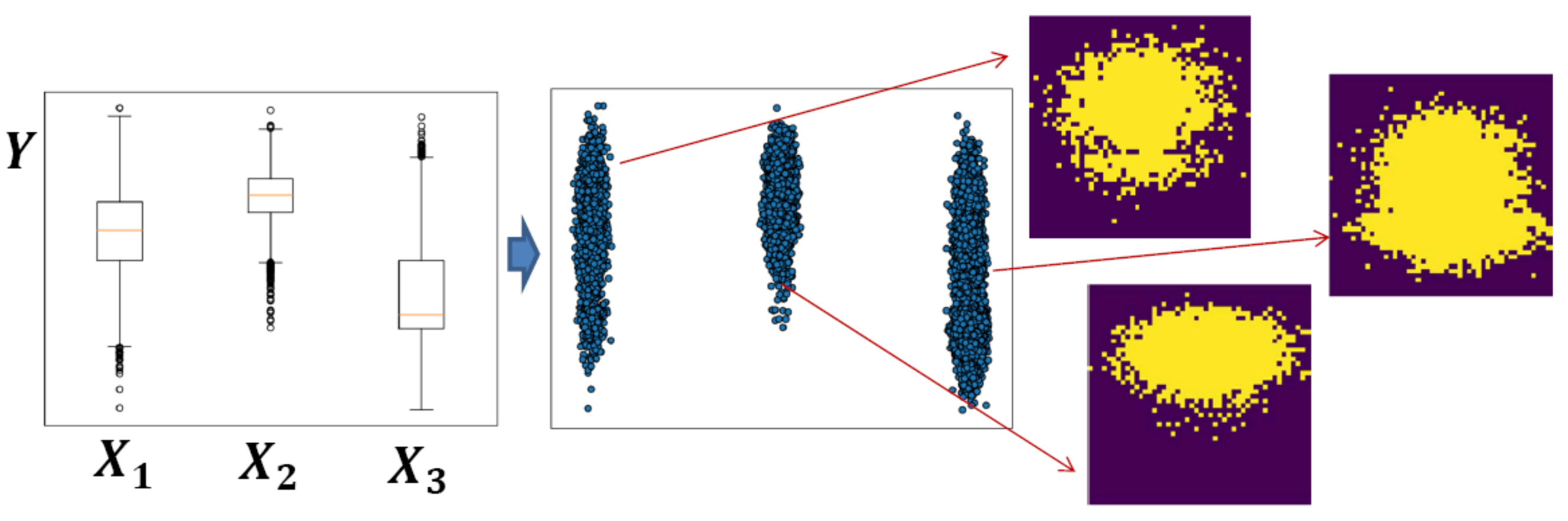}
\vspace{-0.2cm}
\caption{A box plot and their transformed images}
\label{fig15}
\end{figure}

The third type we consider in this work is the box plot. 
Typically, the x-axis of a box plot includes multiple {\it options}. 
The y-axis is a variable of concern. For example, the left
of Figure~\ref{fig15} shows a box plot with three options $X_1, X_2, X_3$.
The red line marks the medium data point. The box denotes the
25\%-75\% quantile range. The two dash lines mark the $\pm$1.5$\times$ the
quantile range. Points outside these dash lines are shown as the outliers. 

A box plot can be used to examine if an issue is with a particular
option. For example, $X_1, X_2, X_3$ can be three tools for the same stage.
The y variable can be the \% of passing dies from a lot. Each data point
represents a lot. What is usually being looked for with a box plot is some
unexpected bias associated with an option, for example higher yield loss is
associated with a particular tool. 

To build a box plot recognizer, a straightforward way is to convert
an entire box plot into a single 48$\times$48 image. 
We found that this approach was not as effective. Instead, 
we convert a box plot into a scatter plot as shown in the middle
of Figure~\ref{fig15}. Then, for each option we convert the spread of
points into a 48$\times$48 image. Therefore, a box plot with
$k$ options would result in $k$ images for the recognizer. The reason
for this choice will be discussed in detail later. 

\vspace{-0.2cm}
\subsection{The recognizer development methodology}
\label{sec04.2}
\vspace{-0.3cm}

As discussed before, an analytic task results in three groups
of plots: non-interesting (usually the majority), known interesting, and novel. 
A novel plot can be deemed interesting or non-interesting by an analyst. 
Our methodology to develop the plot recognizers is trying to develop 
recognizers for non-interesting plots first, followed by recognizers for 
the known interesting plots. Then, the remaining plots
unrecognized by all the recognizers are considered novel. 

There are two main reasons for following this ordering. First, there are
usually many more non-interesting plots than interesting plots. Hence, to begin
with, there are more samples for training a non-interest plot recognizer. 
Second, in practice an analyst might not have a concrete idea what to 
look for in advance. It would be easier to randomly pick a few non-interesting plots
(because there are many) and ask a person to verify their 
non-interest. After majority of the non-interesting plots are filtered out, the remaining
set is much smaller and easier for a person to select those interesting
examples for training an interesting plot recognizer.

\section{Discovering Interesting Wafer Plots}
\label{sec05}
\vspace{-0.3cm}

In this section, the focus is on wafer plots. 
Discussion on correlation plots and box plots will be in the next two
sections. All three sections follow the same methodology described
in Section~\ref{sec04.2} above. 

\vspace{-0.2cm}
\subsection{Developing a sequence of recognizers}
\label{sec05.1}
\vspace{-0.3cm}

Recognizers discussed in this section are developed based on wafer
plots from the product M shown in Table~\ref{table1}. 
The training follows the approach discussed in Section~\ref{sec02}
where five wafer plots are used as the training samples and five wafer
plots are used as the validation samples. 

\begin{figure}[htb]
\centering
\vspace{-0.2cm}
\includegraphics[width=3.3in]{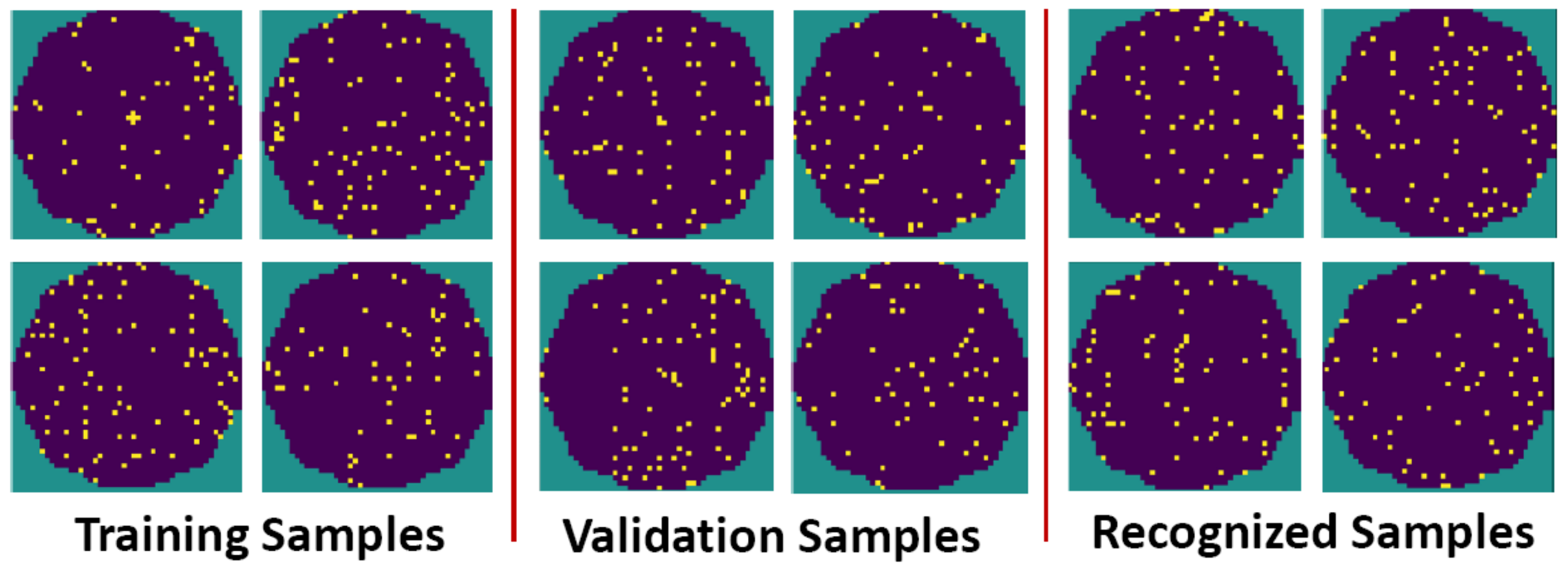}
\vspace{-0.2cm}
\caption{Training, validation, recognized samples for the 1st recognizer}
\label{fig17}
\vspace{-0.2cm}
\end{figure}

Figure~\ref{fig17} shows four of the five training samples and
four of the five validation samples. After the training, the recognizer
is used to scan the rest of the 8300 wafer plots. Four example
recognized plots are also shown in the figure. 

\begin{figure}[htb]
\centering
\vspace{-0.2cm}
\includegraphics[width=3.3in]{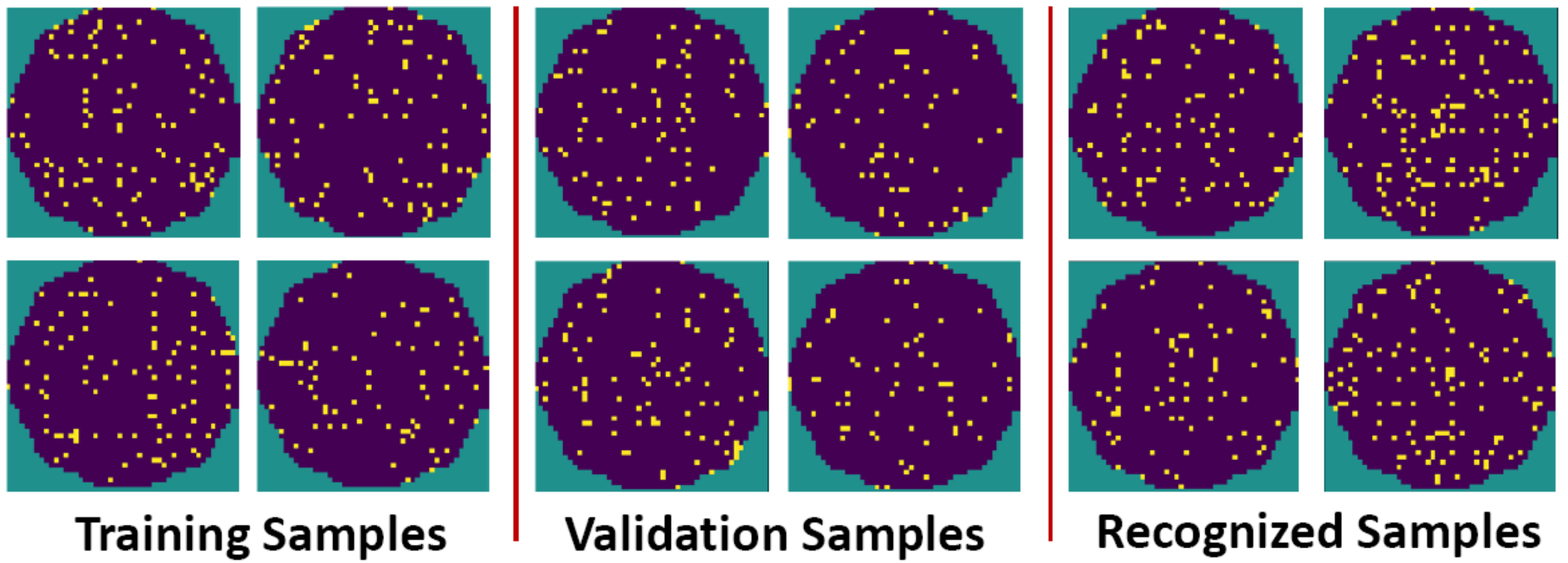}
\vspace{-0.2cm}
\caption{Training, validation, recognized samples for the 2nd recognizer}
\label{fig18}
\vspace{-0.2cm}
\end{figure}

As seen in Figure~\ref{fig17}, the first recognizer is trained to recognize
a random and sparse wafer failing pattern. Majority of the wafer plots are
of this class. 
Then, samples shown in Figure~\ref{fig18} are among those not recognized
by the 1st recognizer. 
The left four are among the five samples used to train the 2nd recognizer.
The middle four are the validation samples. The right four are example
plots recognized by the 2nd recognizer. More than 88\% of the plots are
recognized by the first two recognizers. As a result, less than 12\% of 
the plots remain after applying the first two recognizers. 

Plots recognized by the 2nd recognizer looks similar to those recognized by
the 1st recognizer. However, notice that the 2nd class of plots generally 
contain more failing dies. In theory, the two classes of samples can be
combined to train a single recognizer. However, because of the difference
in their failing density, it requires more samples and would take longer to
converge. We separate them into two recognizers to simplify the training. 

\begin{figure}[htb]
\centering
\vspace{-0.2cm}
\includegraphics[width=3.3in]{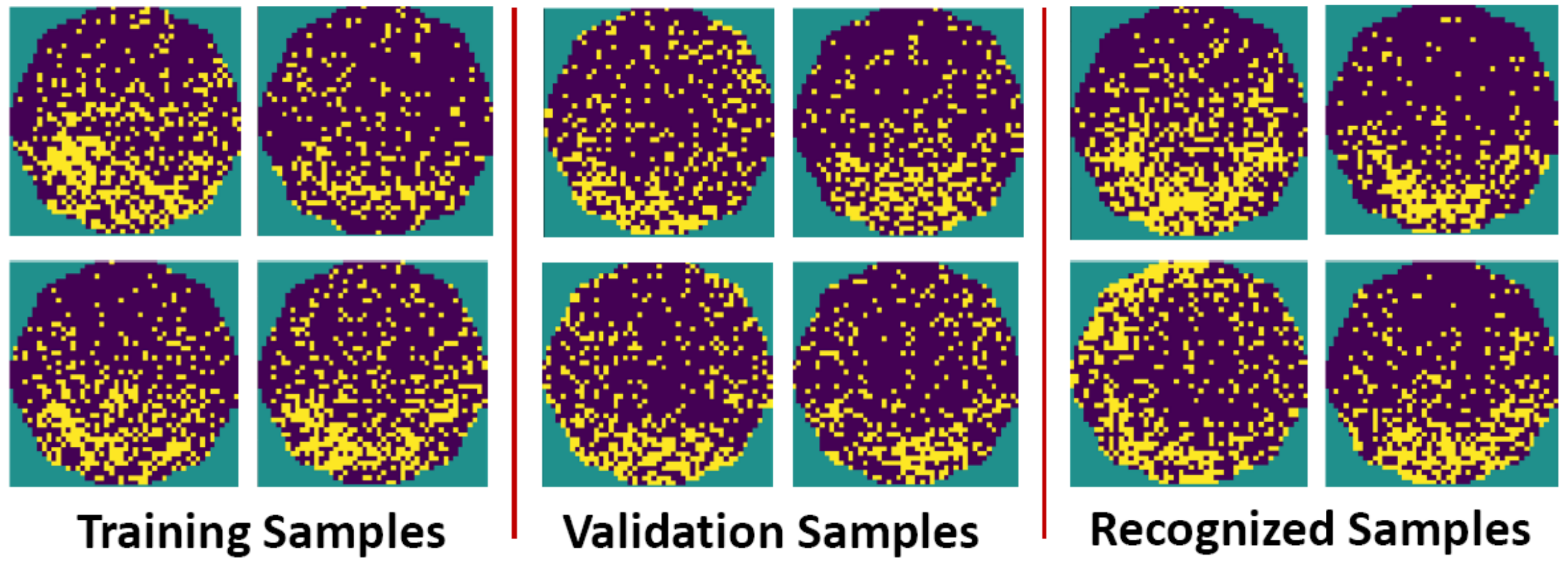}
\vspace{-0.2cm}
\caption{Training, validation, recognized samples for the 3rd recognizer}
\label{fig19}
\end{figure}

Those plots picked up by the first two recognizers are considered non-interesting. 
The 3rd recognizer is trained to recognize a high-density failing pattern
as illustrated in Figure~\ref{fig19}. This class of plots might or might not
be interesting, depending on whether they appear randomly or concentrate
in the same lot. 

\begin{figure}[htb]
\centering
\vspace{-0.2cm}
\includegraphics[width=3.3in]{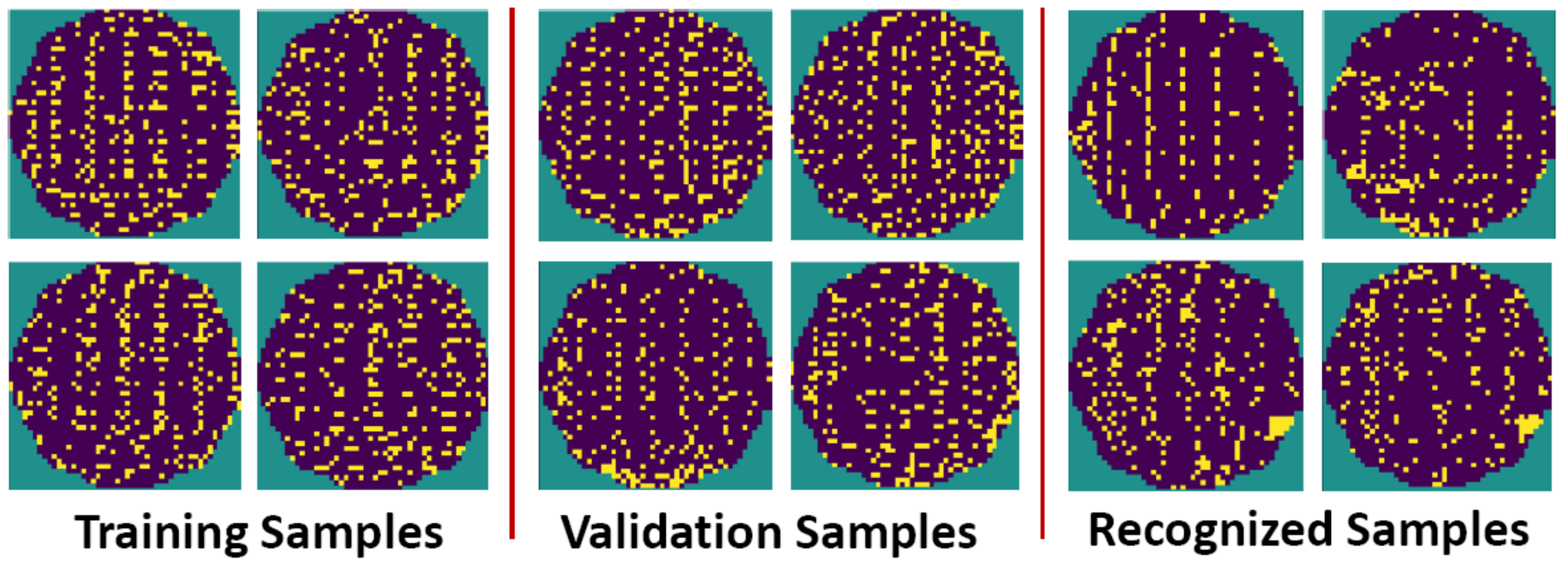}
\vspace{-0.2cm}
\caption{Training, validation, recognized samples for the 4th recognizer}
\label{fig20}
\vspace{-0.2cm}
\end{figure}

The 4th recognizer is trained to recognize a grid pattern as shown in
Figure~\ref{fig20}. This class of plots is interesting and may indicate an
issue in the test probe. There is also a 5th recognizer which is 
the edge pattern recognizer already discussed 
in Section~\ref{sec02} before. 

\vspace{-0.2cm}
\subsection{Six classes of novel plots}
\label{sec05.2}
\vspace{-0.3cm}

\begin{figure}[htb]
\centering
\vspace{-0.2cm}
\includegraphics[width=2.1in]{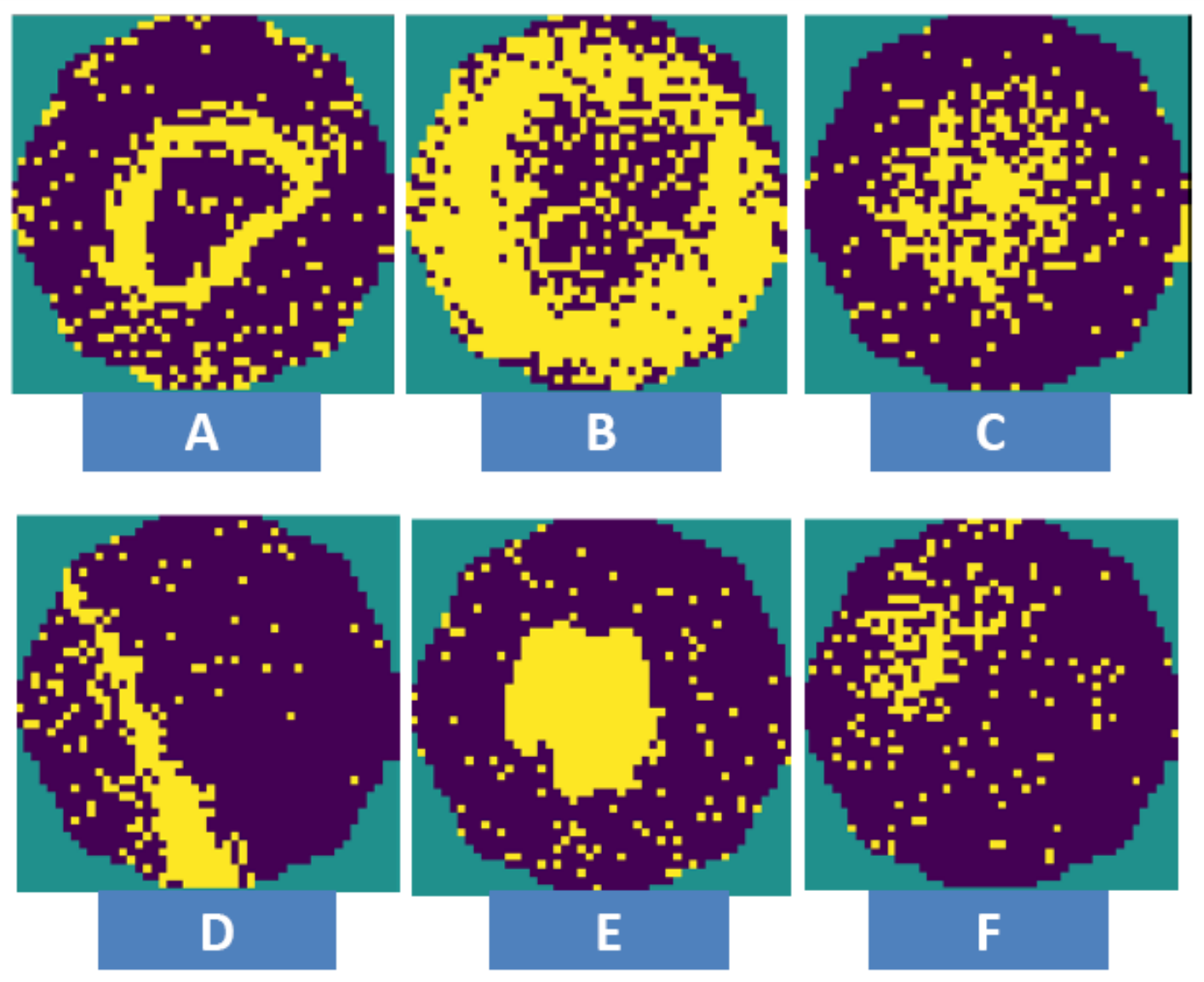}
\vspace{-0.2cm}
\caption{Six novel classes of plots discovered}
\label{fig21}
\vspace{-0.3cm}
\end{figure}

After applying the five recognizers, about 5\% of the plots remain 
unrecognized. They include both non-interesting plots and novel plots. 
About 70\% of these plots can be clustered into one of the
six novel classes as shown in Figure~\ref{fig21}. A class can contain
from at least 10 to over 100 plots and hence, not only the patterns are
novel, but also they appear "systematically." 

\vspace{-0.2cm}
\subsection{Detecting an issue with a production tool}
\label{sec05.3}
\vspace{-0.3cm}

As an example, there are a number of wafer plots sharing the same pattern represented
by the class B example shown in Figure~\ref{fig21}. In fact, this plot is the same
as that presented in Figure~\ref{fig03}-(b) before. Further investigation reveals
that the issue was related to the 3 lift bins used by the Gasonics asher tools. 
Hence, detecting novel patterns as those shown in Figure~\ref{fig21} proved to be
useful in practice. An experienced process engineer could look at a pattern
and start forming guided hypotheses to check the related manufacturing equipments. 
In practice, the recognizers help sort out majority of the wafer plots and bring 
the attention of an engineer onto those novel plots.

\begin{figure}[htb]
\centering
\vspace{-0.2cm}
\includegraphics[width=3.3in]{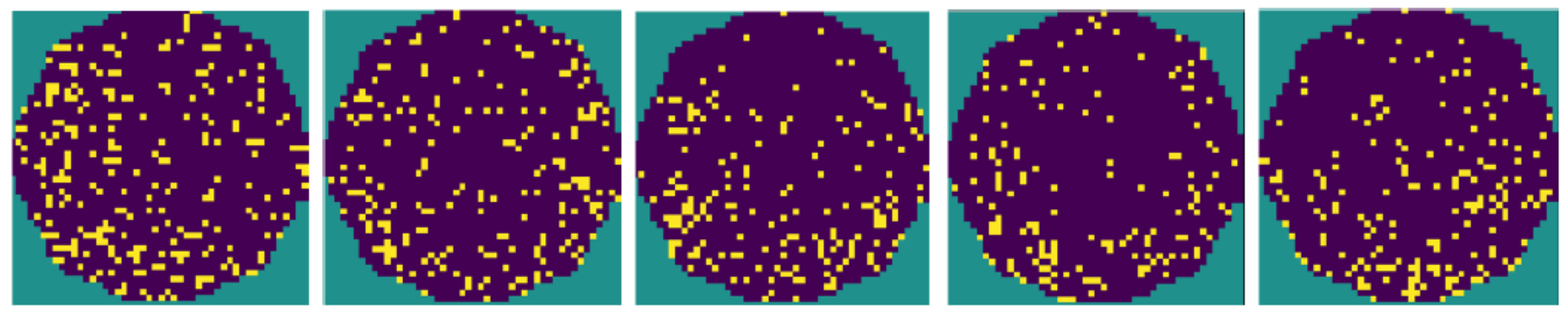}
\vspace{-0.2cm}
\caption{Non-interesting plots missed by the recognizers}
\label{fig22}
\vspace{-0.2cm}
\end{figure}

\vspace{-0.2cm}
\subsection{Non-interesting plots missed by the recognizers}
\label{sec05.4}
\vspace{-0.3cm}

Figure~\ref{fig22} shows some additional examples of the plots 
missed by the five recognizers. These can be deemed as non-interesting plots. 
Notice that the failing locations look also quite random as those plots
shown in Figure~\ref{fig18} but the density of the failing is between
those shown in Figure~\ref{fig18} and those shown in Figure~\ref{fig19}. 
These missing plots suggest that a 6th recognizer can be developed to
recognized a "medium-density random-failing" plot. 
The decision to develop an additional recognizer mostly depends on
the number of plots available for the the training and validation.

\vspace{-0.2cm}
\subsection{Generality of the recognizers}
\label{sec05.5}
\vspace{-0.3cm}

As pointed out in Section~\ref{sec02}, a recognizer developed for a product
line can be applied to the wafer plots from another product line,
even though their numbers of dies on a wafer are quite different.
The five recognizers above are based on product M. Next, we explain
the result by applying these recognizers to wafer plots 
from product A which has about one fifth of the dies on each wafer. 

\begin{figure}[htb]
\centering
\vspace{-0.2cm}
\includegraphics[width=2.1in]{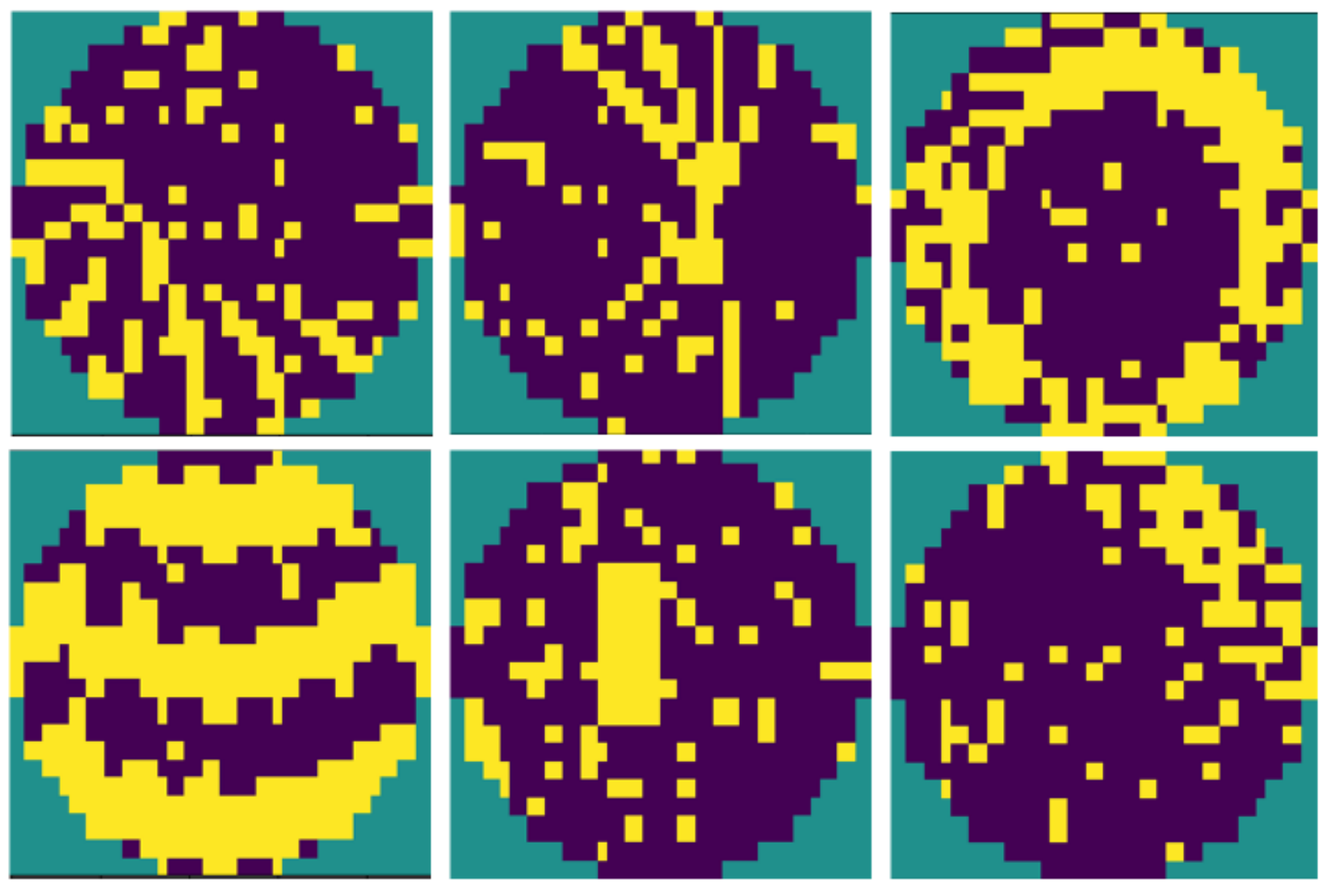}
\vspace{-0.2cm}
\caption{Novel plots discovered on product A}
\vspace{-0.2cm}
\label{fig23}
\end{figure}

The result is that more than 84\% of the plots are recognized by
the first two recognizers, comparing to the 88\% number mentioned
above for product M. As seen, these two numbers are comparable. 
After applying the additional three recognizers, similar to the
result for product M, for product A also about 5\% of the plots 
remain unrecognized. Figure~\ref{fig23} then shows examples of novel
plots found on product A. As seen, these plots show different 
novel patterns than those plots shown in Figure~\ref{fig21} above.

\section{Discovering Interesting Correlation Plots}
\label{sec06}
\vspace{-0.3cm}

Recognizers discussed in this section are developed based on correlation
plots from product A. Again, the development follows the approach 
discussed in Section~\ref{sec02} and transformation from a plot to
an image is discussed with Figure~\ref{fig14} above. 

As shown in Table~\ref{table1}, for product A there are 596 e-tests.
For each plot, we assume that the x-axis is an e-test and 
the y-axis is the number
of fails from a test bin. We select the top 9 bins with the most fails,
which together account for $>$ 90\% of the total fails.
The total number of plots to consider is 5364.
To train a recognizer, we use about 20 training samples and 20
validation samples. 

\begin{figure}[htb]
\centering
\vspace{-0.2cm}
\includegraphics[width=3.3in]{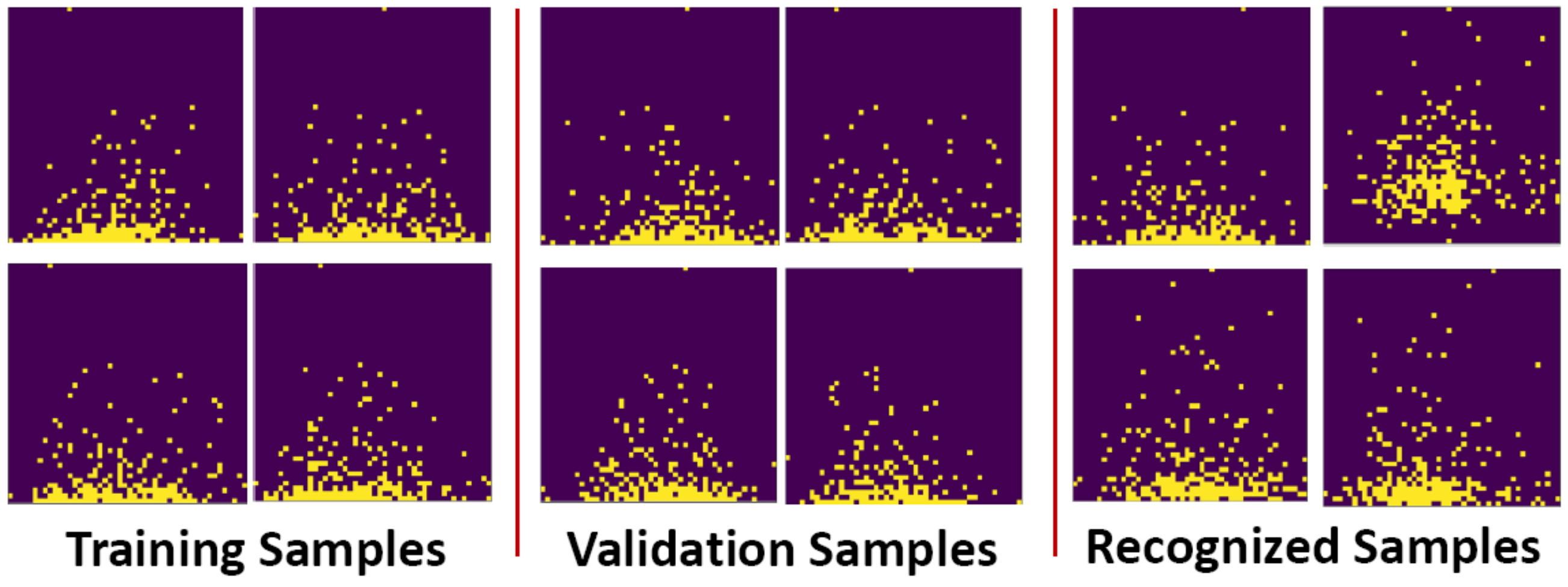}
\vspace{-0.2cm}
\caption{Training, validation, recognized samples for the 1st recognizer}
\label{fig24}
\vspace{-0.2cm}
\end{figure}

Figure~\ref{fig24} shows four of the training samples and
four of the validation samples. After the training, the recognizer
is used to scan the rest of the 5364 correlation plots. Four example
recognized plots are also shown in the figure. 

Plots shown in Figure~\ref{fig24} can be thought of as a typical "no-correlation"
class in this analysis. Each dot represents a lot. The e-test value is the average of
all measured values from all wafers in the lot. 
It can be seen that the e-test values spread randomly across the range shown
with some concentration on the middle of the range. Most of the lots have
few fails and hence, more dots concentrate on the bottom of the image. 
In every plot, there is at least one dot close to the upper edge, i.e. 
the maximum number of fails.  Because of that dot with a large failing
number, the rest of the dots are pushed down in the picture. 

\begin{figure}[htb]
\centering
\vspace{-0.2cm}
\includegraphics[width=3.3in]{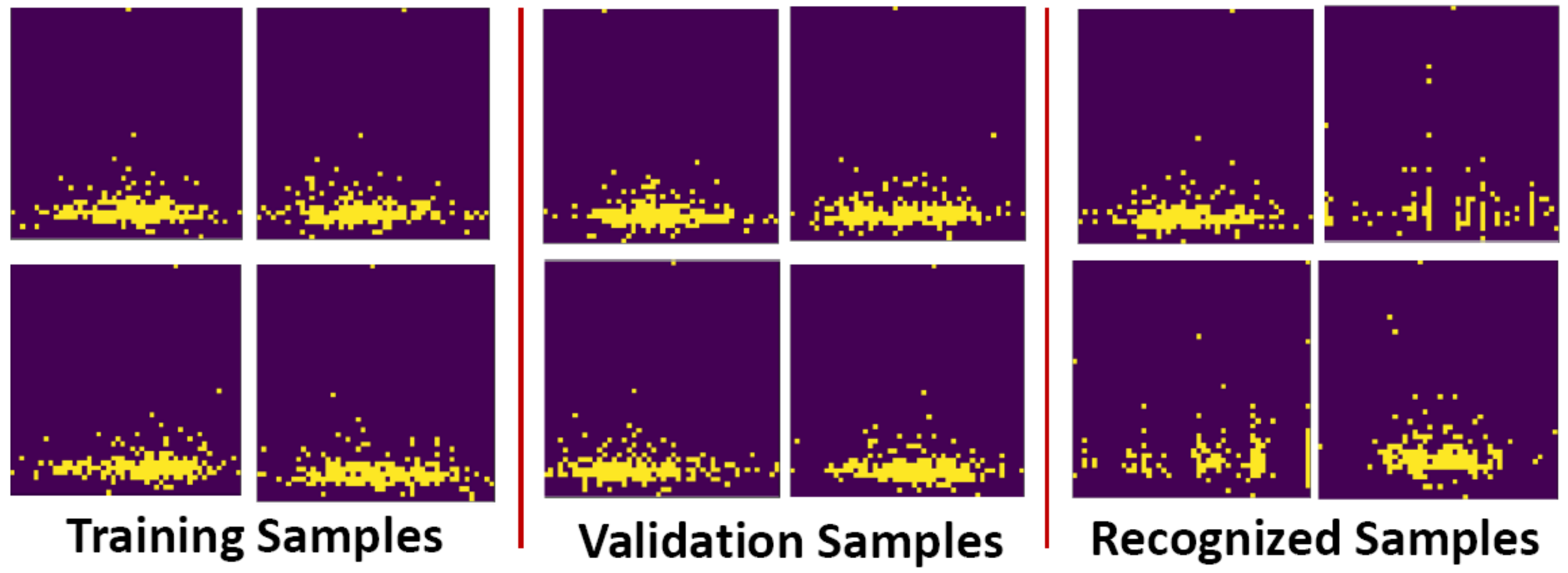}
\vspace{-0.2cm}
\caption{Training, validation, recognized samples for the 2nd recognizer}
\label{fig25}
\vspace{-0.2cm}
\end{figure}

Then, samples shown in Figure~\ref{fig25} are among those not recognized
by the 1st recognizer. Examples of training and validation samples are
shown for training the 2nd recognizer. Again each plot contains a dot close to
the upper edge of the image, resulting in pushing down
further the rest of the dots. Hence, those plots can also be thought of as
another class of "no-correlation."  

The right four are example plots recognized by the 2nd recognizer. Note that
more than 94\% of the plots are recognized by the first two recognizers. 

\vspace{-0.2cm}
\subsection{Two classes of interesting plots}
\label{sec06.1}
\vspace{-0.3cm}

\begin{figure}[htb]
\centering
\vspace{-0.2cm}
\includegraphics[width=3.3in]{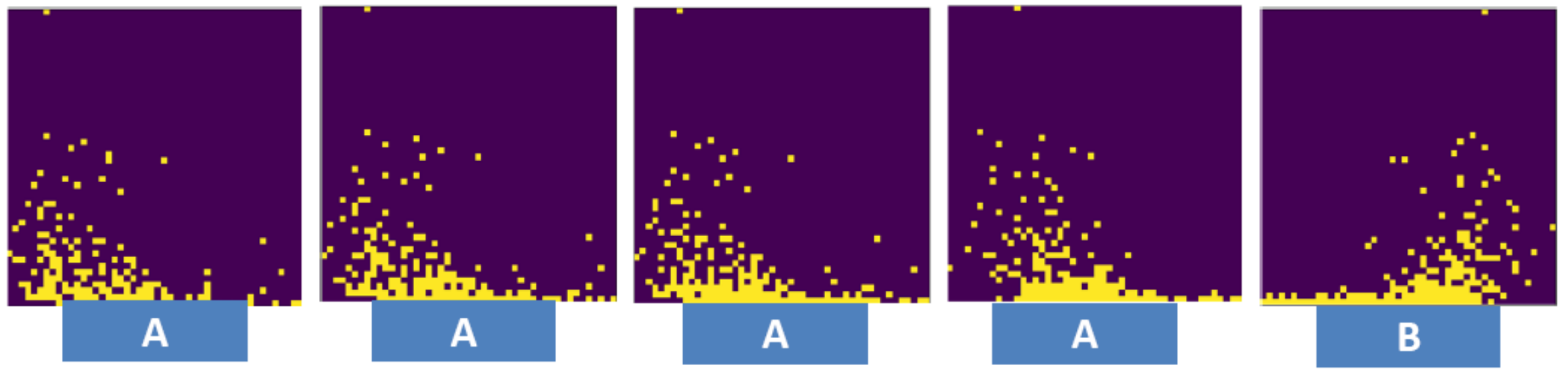}
\vspace{-0.2cm}
\caption{Two classes of interesting plots discovered}
\label{fig26}
\vspace{-0.2cm}
\end{figure}

Among the remaining plots, two classes of interesting plots are found as
illustrated in Figure~\ref{fig26}. Class A shows that a smaller e-test
value tends to have more fails. Class B shows that a larger
e-test value tends to have more fails. As explained with Figure~\ref{fig03}-(a)
before, the correlation coefficients with such plots are small. Hence,
a search constrained by a high correlation coefficient would not have 
found these interesting plots. 

\vspace{-0.2cm}
\subsection{An application scenario for improving yield}
\label{sec06.2}
\vspace{-0.2cm}

\begin{figure}[htb]
\centering
\vspace{-0.2cm}
\includegraphics[width=2.3in]{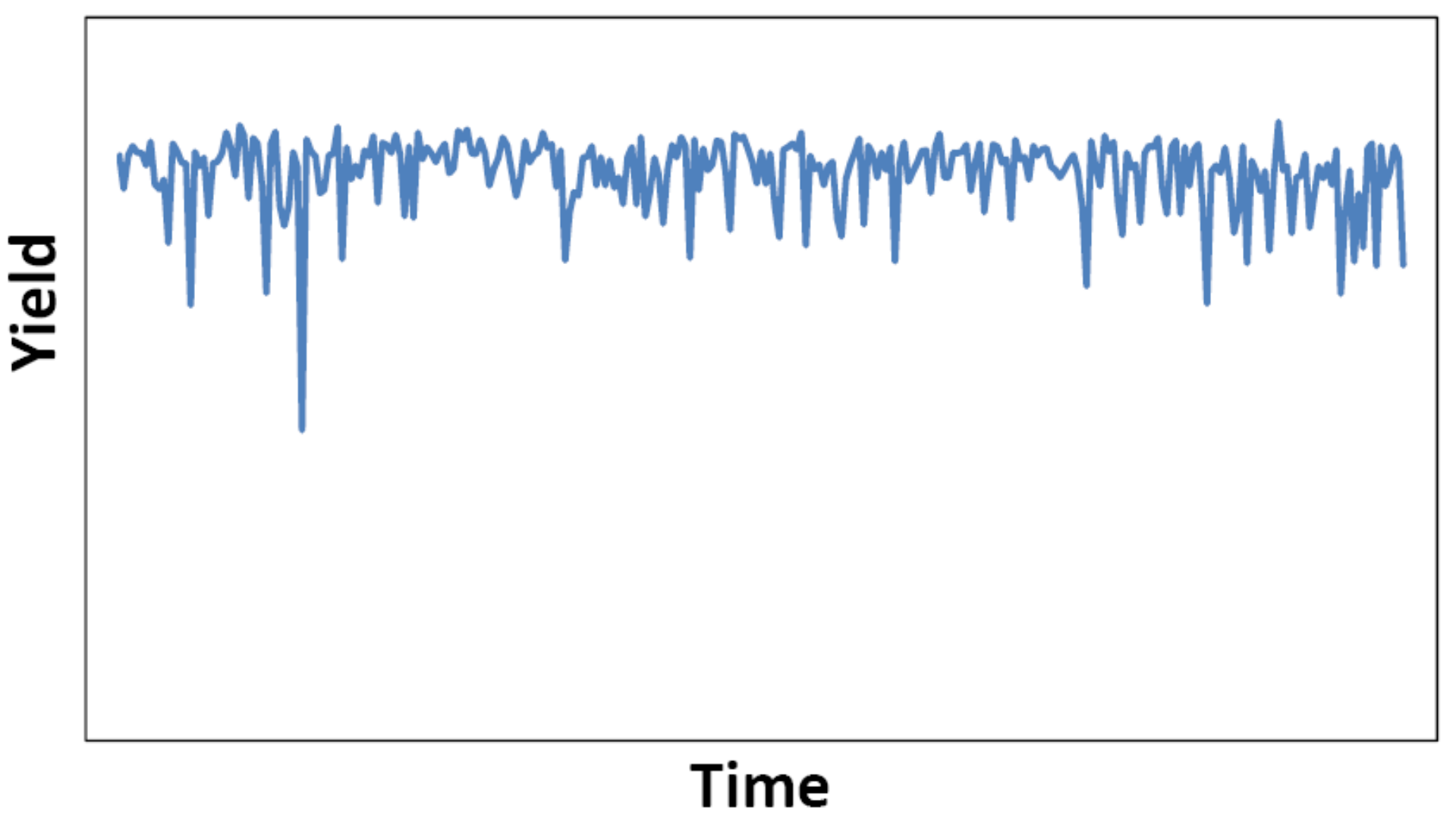}
\vspace{-0.2cm}
\caption{Yield fluctuation over time on product A}
\vspace{-0.2cm}
\label{fig28}
\end{figure}

Figure~\ref{fig28} illustrates an application scenario. The figure
shows the lot-based yield fluctuation over time. As seen, some lots
have noticeably lower yield than others. 

One analytic task is to find an e-test parameter that correlates
to the number of fails from a test bin. The correlation plots used
in the experiment above were generated for this task. 
Therefore, e-test parameters found with the plots in Figure~\ref{fig26} 
are all candidates for further analysis. 

After further investigation based on the e-test parameter with 
the class B plot and the e-test parameter with one of the class A plot,
it was found that after a change of recipe in a tool, the class B
parameter drifted toward larger values and the class A parameter 
drifted toward smaller values. After the recipe was rolled back to the 
previous version, those drifts disappeared and the yield was improved. 

\vspace{-0.2cm}
\subsection{Plots missed by the recognizers}
\label{sec06.3}
\vspace{-0.3cm}

\begin{figure}[htb]
\centering
\vspace{-0.2cm}
\includegraphics[width=3.3in]{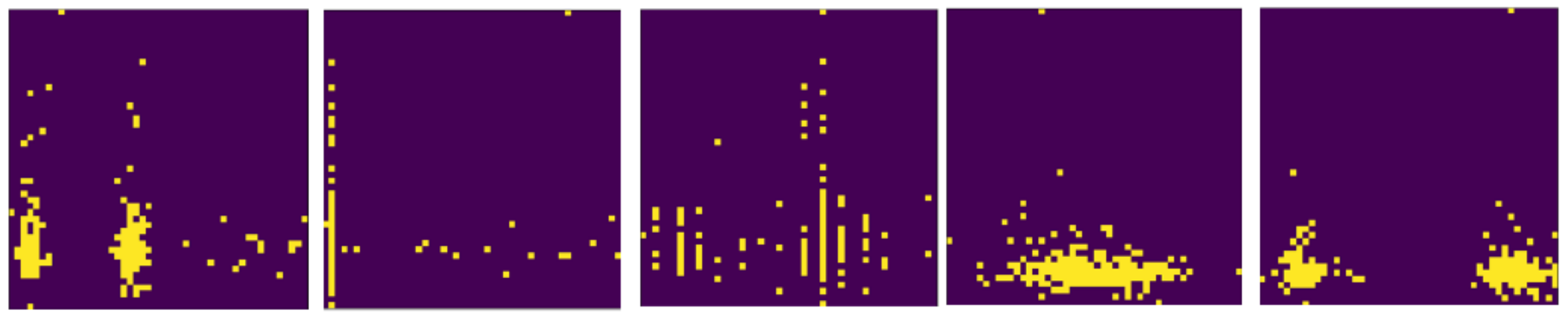}
\vspace{-0.2cm}
\caption{Examples of other plots missed by the two recognizers}
\label{fig27}
\vspace{-0.3cm}
\end{figure}

There are non-interesting plots missed by the two recognizers. 
Figure~\ref{fig27} shows some examples. Those plots are not interesting
because they reveal little correlation between the e-test value
and the number of fails. 

\vspace{-0.2cm}
\subsection{An adversarial example}
\label{sec06.4}
\vspace{-0.3cm}

The recognizers above are developed based on the definition that
y-axis is the number of fails in a test bin. If we change this
definition to be the test values from a test and produce new plots,
then an interesting question would be whether
or not the above recognizers can still be used to filter the 
non-interesting plots with the new y-axis definition. 

\begin{figure}[htb]
\centering
\vspace{-0.2cm}
\includegraphics[width=3.3in]{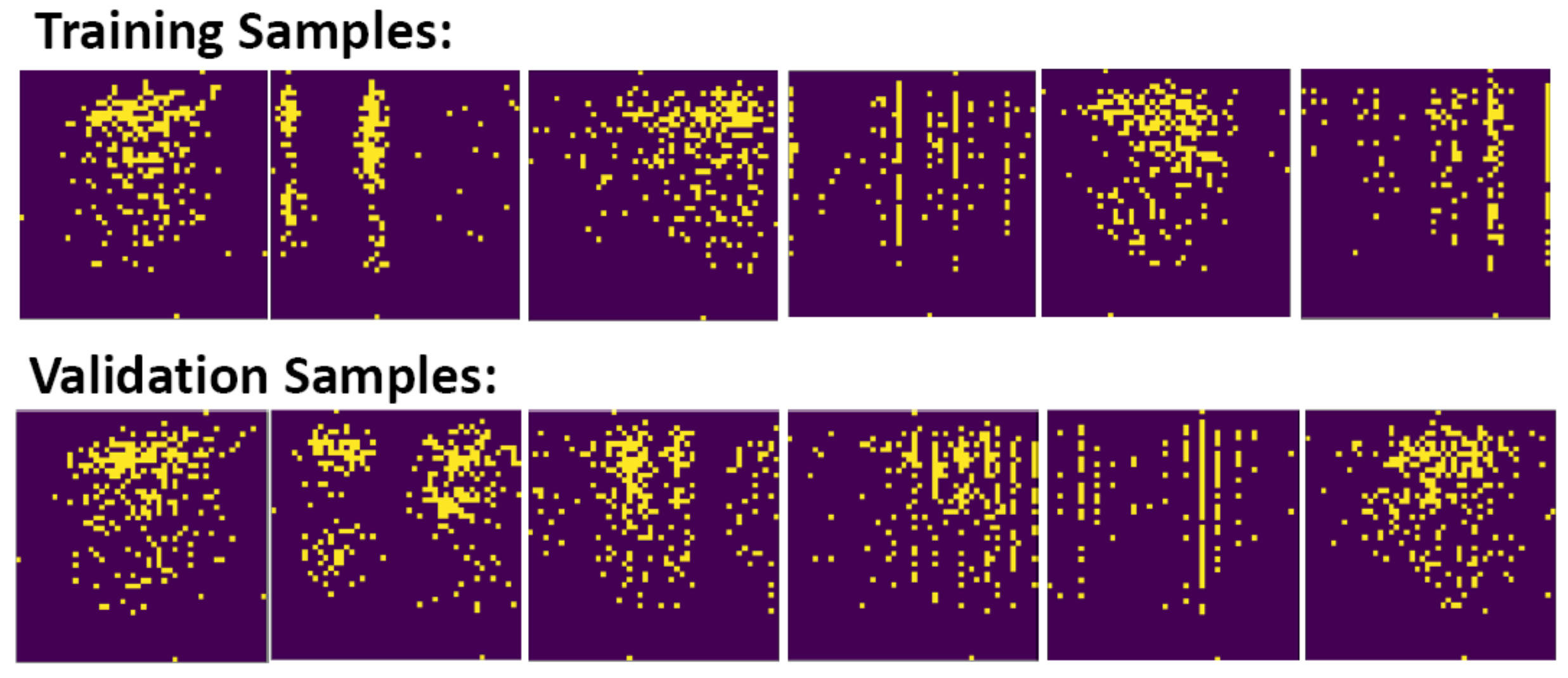}
\vspace{-0.2cm}
\caption{Examples of training/validations samples for the recognizer}
\label{fig29}
\vspace{-0.2cm}
\end{figure}

Based on five selected tests and the 596 e-tests, 2908 new plots
are generated. When the recognizers are applied to those new plots,
most of them are not recognized. Figure~\ref{fig29} shows some
examples of those unrecognized plots. As it can be seen, the new
plots look quite differently from those shown in Figure~\ref{fig24}
and Figure~\ref{fig25} above. The patterns in those new plots
appear to be more diverse and random. 

Suppose we train a new recognizer specific to the new plots. 
Figure~\ref{fig29} shows some of the training and validation
samples. We use 100 training and 100 validation samples. These
numbers are larger than before because there is no clear systematic
patterns observed on the new plots. Hence, we expect it is more
difficult to train a recognizer. 

\begin{figure}[htb]
\centering
\vspace{-0.2cm}
\includegraphics[width=3.3in]{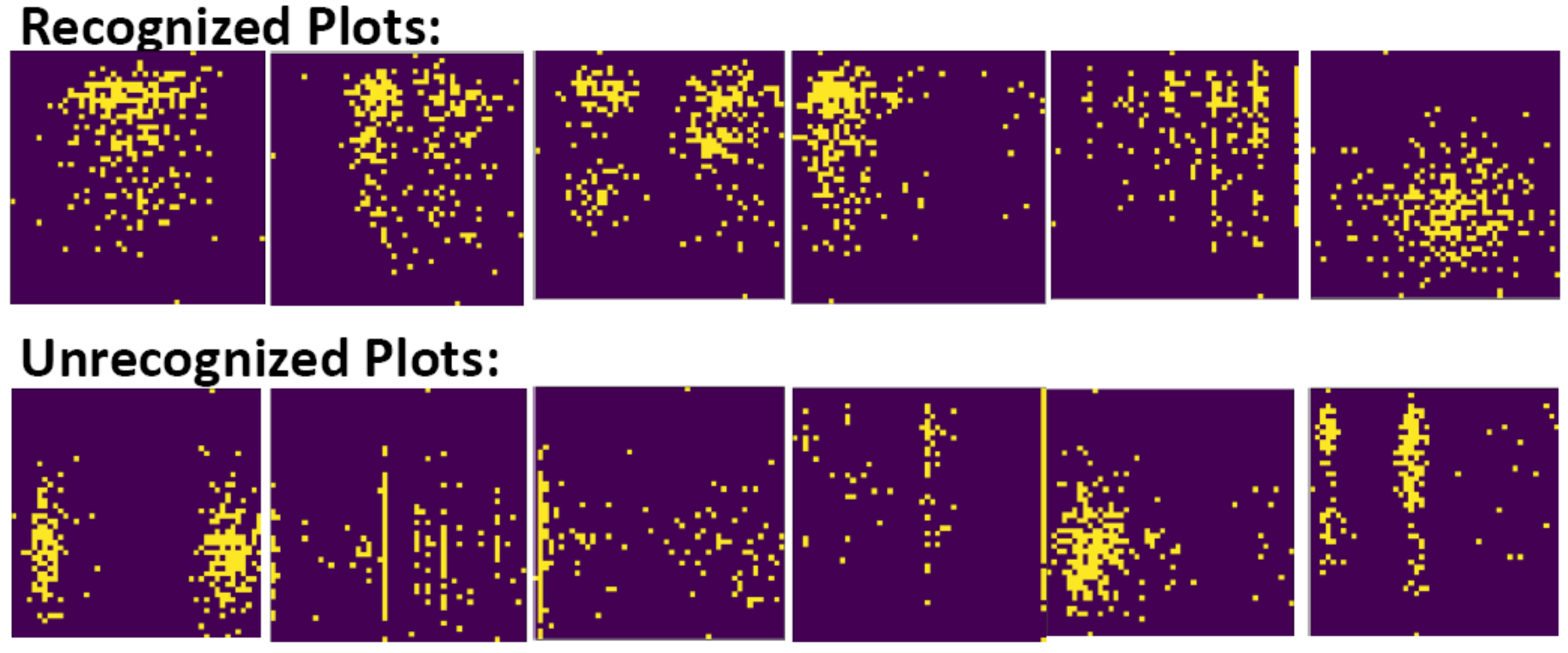}
\vspace{-0.2cm}
\caption{Recognized and unrecognized example plots by the recognizer}
\label{fig30}
\vspace{-0.2cm}
\end{figure}

After the training (with 5 hours training time), most of the plots are 
recognized with only 110 plots
remaining. Figure~\ref{fig30} shows examples of the recognized and
unrecognized plots. For some recognized plots, it appears that we can
find a similar training or validation plot in Figure~\ref{fig29} to 
explain why they are recognized. However, the 
rightmost unrecognized plot also looks the same as the the 2nd training
sample from the left (they are actually slightly different). 

The rightmost unrecognized plot is called an {\it adversarial example} \cite{AS2013},
a slightly perturbed example that can fool a neural network (NN) model. 
This is a well-known issue concerning the robustness of a NN model \cite{AS2013}. 
For training other recognizers before, the training plots share some similar
features and hence, we did not observe such an adversarial example. 
In contrast, the samples in Figure~\ref{fig29} are much more diverse and random. 
As a result, the recognizer is less robust and it is easier for adversarial
examples to exist. 
In this work, we acknowledge this well-known issue in NN \cite{AS2013} but 
will leave it to future research. 

\section{Box-Plot Based Recognizers}
\label{sec07}
\vspace{-0.2cm}

When a box plot is used by a person, the plot usually includes a small group of options 
for the convenience of visualization. However, for a recognizer it can process
all box plots together as long as they have the same y-axis definition. 
For example, suppose y-axis is the yield. Then, if an option shows no
bias, the vertical distribution should look similar to the original
yield distribution. Hence, regardless of what each option means, as
long as the y-axis is fixed, many of their vertical distributions 
should look similar. This is why in our methodology, we follow the idea
explained in Figure~\ref{fig15} before. In this way, each option is associated
with an image. After the transformation, all options are considered collectively 
by a recognizer to recognize the "normal behavior." An unrecognized image
then tells that the corresponding option behaves differently from others.

\vspace{-0.2cm}
\subsection{Monitoring production tools}
\label{sec07.2}
\vspace{-0.3cm}

To show how such a "box-plot" recognizer can be useful, we apply the idea
to monitor production tools for product A (because of the yield fluctuation
issue shown in Figure~\ref{fig28}). The goal is to detect if any tool behaves
unexpectedly as comparing to other tools for the same stage. In the
production process, there are more than 790 stages with two or
more tools. For each tool, an image is extracted and yield distribution
is represented as the spread vertically (i.e. y-axis). The horizontal
spread is artificially randomized and has no particular meaning. Each dot 
represents a lot. 

\begin{figure}[htb]
\centering
\vspace{-0.2cm}
\includegraphics[width=3.3in]{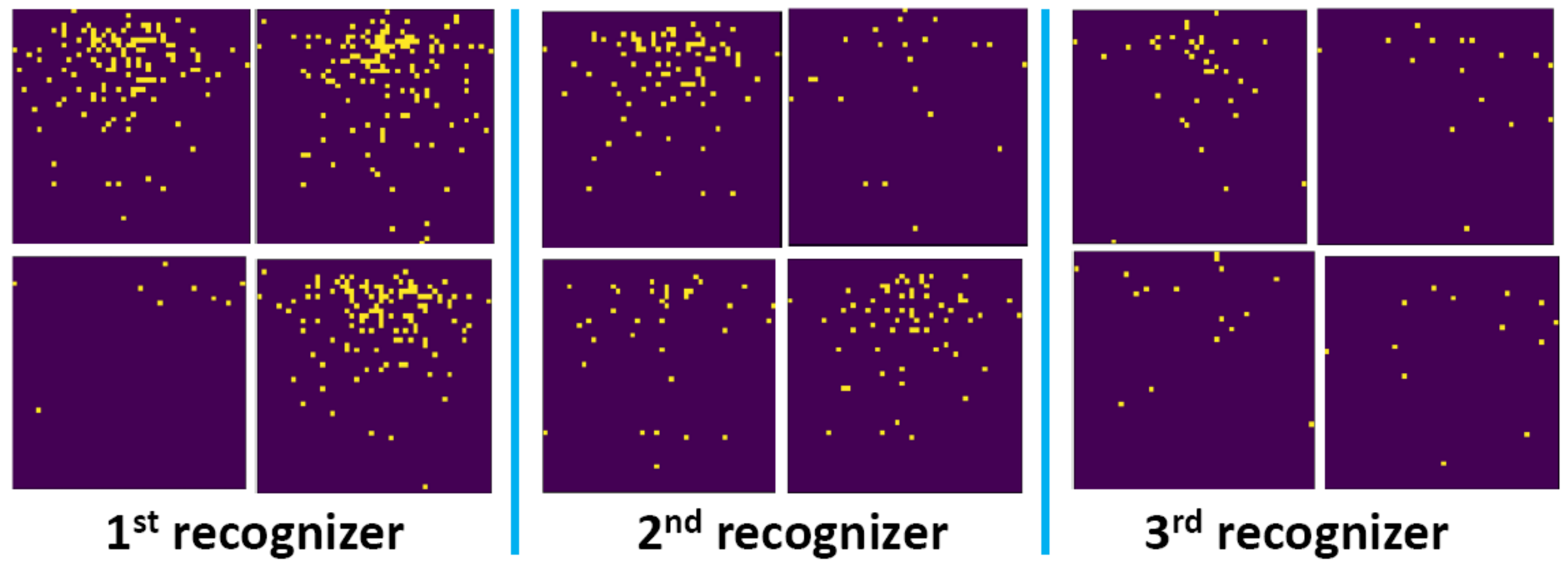}
\vspace{-0.2cm}
\caption{Examples of training samples for the three recognizers}
\label{fig34}
\end{figure}

\begin{figure}[htb]
\centering
\vspace{-0.2cm}
\includegraphics[width=3in]{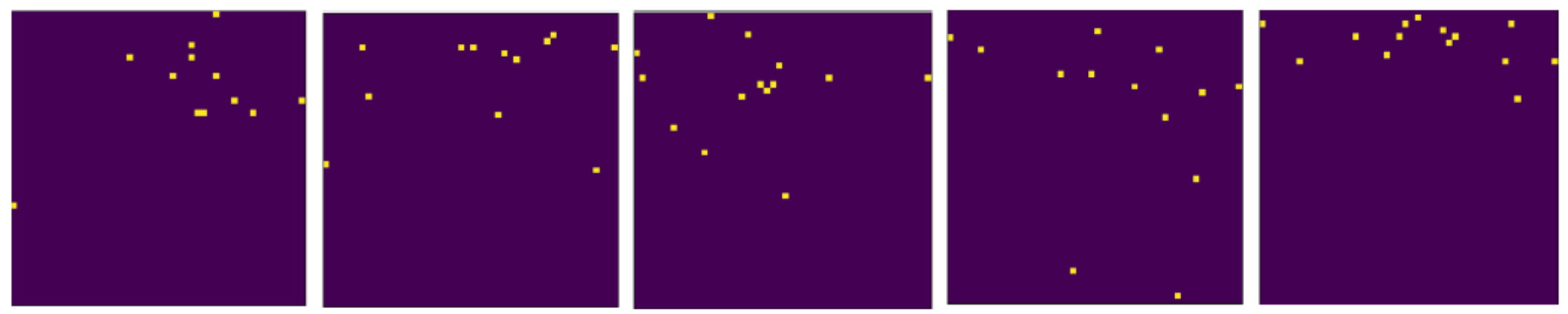}
\vspace{-0.2cm}
\caption{Unrecognized images each corresponding to a tool}
\label{fig35}
\vspace{-0.2cm}
\end{figure}

Three recognizers are trained in sequence following the same
methodology used above. Figure~\ref{fig34} shows some
training samples for training the recognizers. For training one
recognizer, 20-40 samples are used. After applying the three recognizers, only 29 plots are
left. Figure~\ref{fig35} shows examples of those plots. 
 
Figure~\ref{fig34} tells that most of the tools follow three classes
of typical "behavior" in term of the resulting yield (i.e. recognized
by the three recognizers). 
Then, the remaining images in Figure~\ref{fig35} tells that those
corresponding tools are used by much fewer lots and hence, their
images look different. Together, the result show that no tool
has a strong bias in terms of the yield. Note that the 
same approach can be applied to monitor chambers and other 
process options. 

\vspace{-0.2cm}
\subsection{Monitoring testers}
\label{sec07.1}
\vspace{-0.2cm}

\begin{figure}[htb]
\centering
\vspace{-0.2cm}
\includegraphics[width=3.3in]{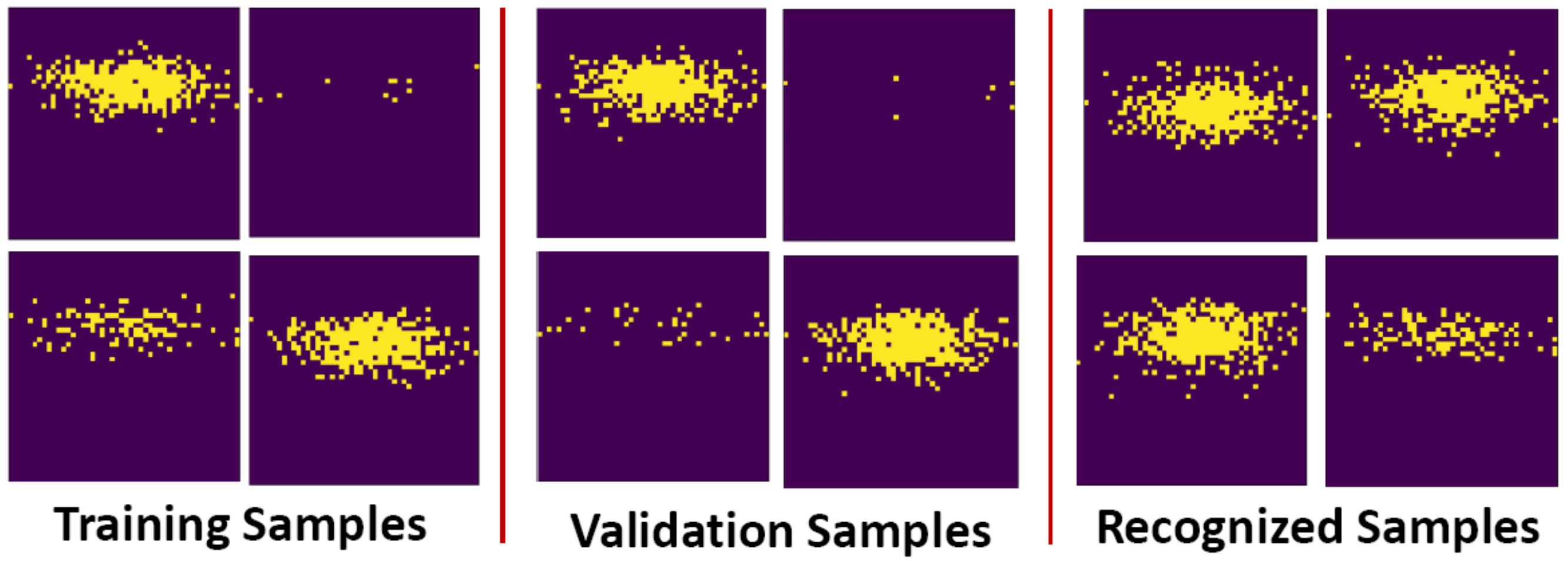}
\vspace{-0.2cm}
\caption{Training, validation, recognized samples for the one recognizer}
\label{fig32}
\end{figure}

The approach can be applied to monitor testers and compare their
statistical behaviors. For example, in Figure~\ref{fig32}, each
image represents a wafer. A dot represents a part from the wafer.
The y-axis is the test value of a final test, measured on tester\#1. 
This is also based on product A.  
One recognizer is trained for tester\#1, with 20 training samples
and 20 validation samples. Figure~\ref{fig32} shows some examples. 
The recognizer recognizes all plots derived from tester\#1 
except for 32 plots deemed non-interesting after inspection. 

\begin{figure}[htb]
\centering
\vspace{-0.2cm}
\includegraphics[width=2.9in]{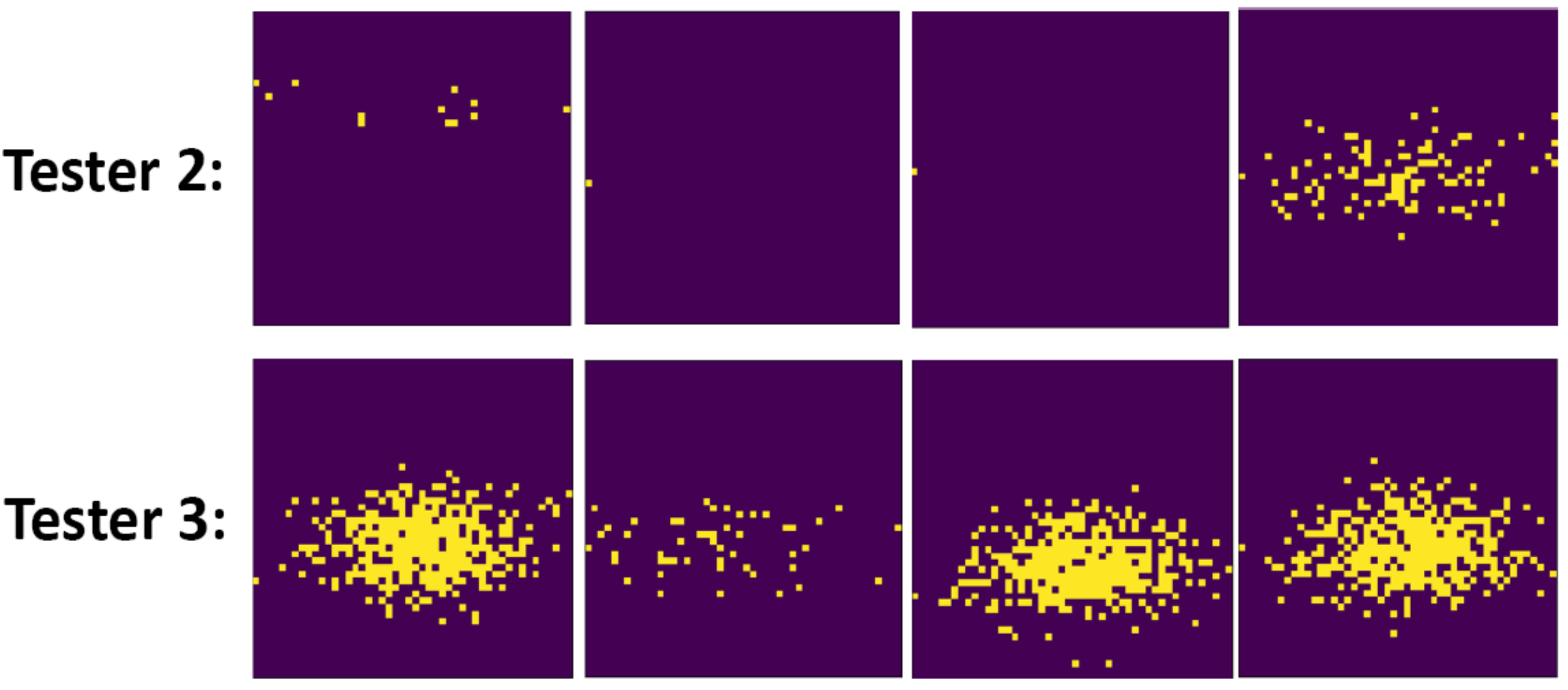}
\vspace{-0.2cm}
\caption{Unrecognized images from the other two testers}
\label{fig33}
\vspace{-0.2cm}
\end{figure}

Similar plots are obtained for tester\#2 and tester\#3. The recognizer
is applied on those plots. For tester\#2, all plots are recognized
except four of them. These four plots are shown in Figure~\ref{fig33}.
Because most of the plots from tester\#2 are recognized, we can say
that tester\#1 and tester\#2 behaves "statistically"' similarly 
(and also across time because
different wafers of parts can be tested at different times). 

For tester\#3, the situation is quite different. More than 50\% of
the plots are not recognized. Some examples are also shown in Figure~\ref{fig33}.
A careful look on those plots can tell the reason why they are not
recognized - the dots tend to be vertically lower and have a wider
spread than those shown in Figure~\ref{fig32}. This suggests that tester\#3 
behaves differently from the other two testers. Further investigation 
confirms that tester\#3 does have six times more failing parts (in the test
bin containing the final test) than that from the other two testers combined. 
This signals an issue with tester\#3. After the issue was resolved,
the yield was improved. 

\section{Conclusion}
\label{sec08}
\vspace{-0.2cm}

In this work, we use two deep CNNs to implement GANs for training a plot recognizer.
Multiple recognizers can be trained for a particular type of analytics in order to recognize
both non-interesting and known interesting plots in the respective application context. 
Then, unrecognized
plots are novel and their interests can be decided manually. We consider three
types of plots commonly used in yield data analytics: the wafer plot,
the correlation plot, and the box plot. We use data collected from two product
lines to illustrate the development of various plots recognizers. 
We discuss four application scenarios to
explain their usefulness in practice where in three scenarios engineers were
able to improve the yield based on the findings, and in one scenario the
plot recognizers help ensure that no individual tool as a whole causes a
significant shift in the yield. 

The proposed plot recognizer based methodology is general and can be applied to 
other types of plots and in other application scenarios. However, the effectiveness
and the generality of a recognizer can largely depend on the classes of images to be 
trained with. If a single recognizer is trained to recognize images with very diverse 
and/or random features, this might demand a large training set, a carefully-selected 
validation set, and/or a longer training time. This might also 
cause difficulty to converge in training or result in a model of which robustness
can be in question (e.g. the adversarial example discussed in Section~\ref{sec06.4}). 
As a result, training such a recognizer might become impractical. Further study 
is required to understand the limitations of the methodology and the scope of 
its applicability in practice.



\bibliographystyle{IEEEtran}
\itemsep=0pt
\topsep = 0pt
\partopsep=0pt
\itemsep= 0cm
\parsep=0pt
\parskip= 0cm
\bibliography{MyBib}

\end{document}